\documentclass[10pt,journal,compsoc]{IEEEtran}

\usepackage{booktabs}
\usepackage{makecell}
\usepackage{multirow}
\usepackage{graphicx}
\usepackage{color}
\usepackage{hyperref}
\usepackage{ulem}
\usepackage{capt-of}
\usepackage[table]{xcolor}
\usepackage{graphicx}

\ifCLASSOPTIONcompsoc
  \usepackage[nocompress]{cite}
\else
  \usepackage{cite}
\fi

\begin{document}

\title{RealLiFe: Real-Time Light Field Reconstruction via Hierarchical Sparse Gradient Descent}
\author{Yijie~Deng$^*$,
        Lei~Han$^*$,
        Tianpeng~Lin,
        Lin~Li,
        Jinzhi~Zhang,
        and~Lu~Fang$^{\S}$

    \thanks{
        Yijie Deng, Jinzhi Zhang and Lu FANG are with Dept. of Electrical Engineering at Tsinghua University and also Beijing National Research Center for Information Science and Technology (BNRist), Beijing 100084. Yijie Deng and Jinzhi Zhang are also with Tsinghua Shenzhen International Graduate School. Lei Han, Tianpeng Lin and Lin Li are with Huawei Technology. 
    \newline
        $*$: Equal Contribution. 
    \newline
        $\S$: Correspondence Author 
            (\href{mailto:fanglu@tsinghua.edu.cn}{fanglu@tsinghua.edu.cn},
            \url{http://luvision.net}).
    }
}

\IEEEtitleabstractindextext{%

\begin{abstract}
With the rise of Extended Reality (XR) technology, there is a growing need for real-time light field reconstruction from sparse view inputs. 
Existing methods can be classified into offline techniques, which can generate high-quality novel views but at the cost of long inference/training time, and online methods, which either lack generalizability or produce unsatisfactory results. 
However, we have observed that the intrinsic sparse manifold of Multi-plane Images (MPI) enables a significant acceleration of light field reconstruction while maintaining rendering quality.
Based on this insight, we introduce \textbf{RealLiFe}, a novel light field optimization method, which leverages the proposed Hierarchical Sparse Gradient Descent (HSGD) to produce high-quality light fields from sparse input images in real time.
Technically, the coarse MPI of a scene is first generated using a 3D CNN, and it is further optimized leveraging only the scene content aligned sparse MPI gradients in a few iterations.
Extensive experiments demonstrate that our method achieves comparable visual quality while being 100x faster on average than state-of-the-art offline methods and delivers better performance (about 2 dB higher in PSNR) compared to other online approaches.
    
\end{abstract}

}

\maketitle
\IEEEdisplaynontitleabstractindextext
\IEEEpeerreviewmaketitle

\IEEEraisesectionheading{\section{Introduction}\label{sec:introduction}}
\IEEEPARstart{W}{ith} recent advances in extended reality (XR) and autostereoscopic 3D display technologies, there is an urgent demand for real-time light field reconstruction systems capable of generating high-quality novel views from sparse input images. This technological imperative drives critical applications including low-latency 3D telepresence systems, interactive augmented reality interfaces, and immersive virtual reality platforms with adaptive viewpoint rendering capabilities.

Despite significant progress in neural light field reconstruction \cite{Flynn2019DeepViewVS,Broxton2020ImmersiveLF} and neural radiance field representations \cite{Mildenhall2020NeRFRS,Barron2021MipNeRFAM}, achieving photorealistic rendering quality while maintaining real-time constraints for interactive applications remains challenging. Current approaches face fundamental trade-offs between visual performance and computational efficiency.

\begin{figure}[!t]
    \centering
    \includegraphics[width=\columnwidth]{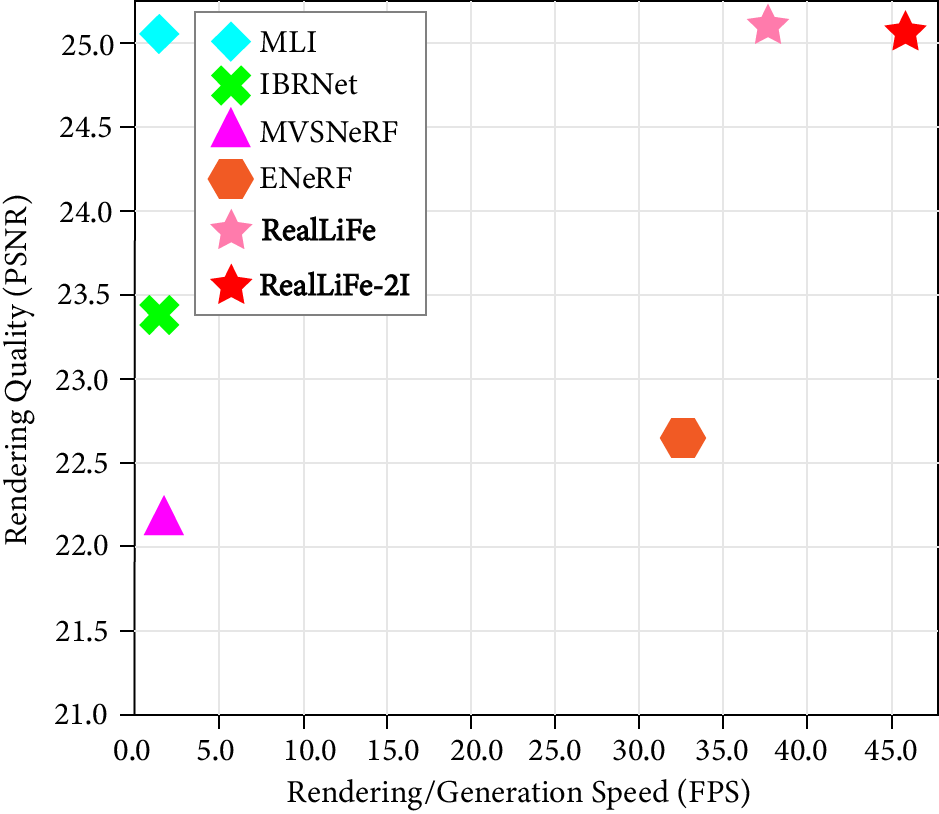}
    \caption{Rendering quality and efficiency comparison with state-of-the-art novel view synthesis methods\cite{Wang2021IBRNetLM,Chen2021MVSNeRFFG,Lin2021EfficientNR} and light field reconstruction methods\cite{Solovev2022SelfimprovingMI} on Real Forward-Facing\cite{Mildenhall2019LocalLF} of image size $512\times 384$. \textbf{RealLiFe} is our default model with 3 iterations of gradient descent, and \textbf{RealLiFe-2I} is one with 2 iterations of gradient descent. }
    \label{fig_comparison}
\end{figure}

\begin{figure*}[!t]
    \centering
    \includegraphics[width=\textwidth]{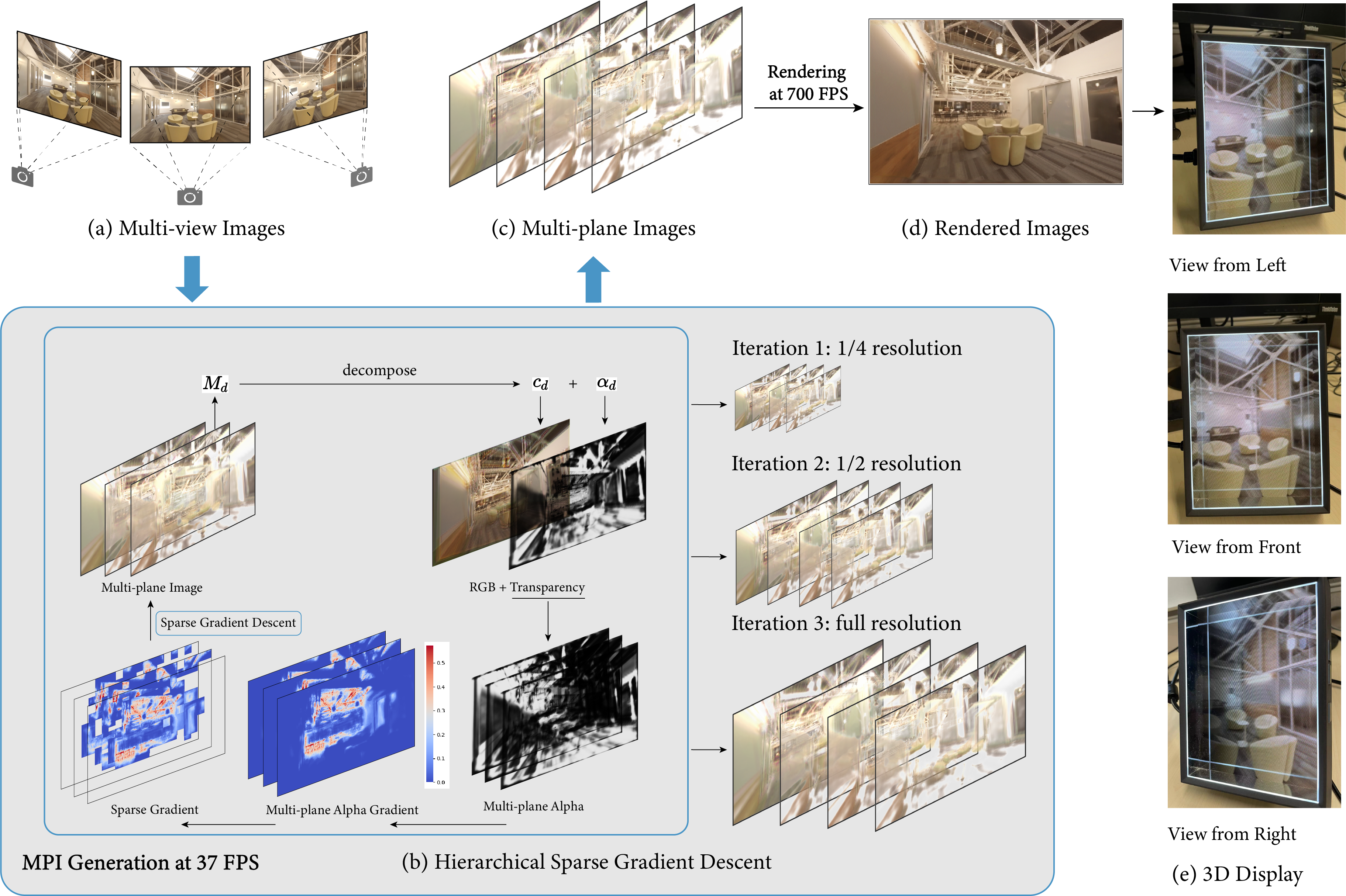}
    \caption{Application of our method to support a real-time 3D display. (a) Sparse multi-view images that serve as the input to our model. 
    (b) Our Hierarchical Sparse Gradient Descent (HSGD) is capable of generating multi-plane images (c) online at around 35 FPS. (d) Novel views can then be rendered offline from a Multi-plane Image at approximately 700 FPS.
    (e) Several novel views rendered from the MPI provide enough light field information to a 3D display, specifically a looking glass that supports naked-eye 3D effects from a wide range of views. (GD is short for gradient descent.) }  
    \label{fig_teaser}
\end{figure*}

To enable live light field reconstruction, the supporting algorithms and representations should be capable of producing high-quality view synthesis results in real time and generalize well to unseen scenes. However, current approaches face difficulties in achieving a balance between visual performance and real-time effectiveness.

\begin{itemize}
\item \textbf{Implicit representation methods}, such as NeRF \cite{Mildenhall2020NeRFRS} and its derivatives \cite{Lindell2020AutoIntAI,Rebain2020DeRFDR,Reiser2021KiloNeRFSU,Garbin2021FastNeRFHN}, produce high-quality view synthesis but require extensive per-scene optimization and dense view inputs. While generalization techniques \cite{Liu2022NeuralRF,Trevithick2020GRFLA,Yu2021pixelNeRFNR,Wang2021IBRNetLM,Chen2021MVSNeRFFG,Chibane2021StereoRF} can generate generalizable novel views using multi-view stereo geometry rules, they remain computationally intensive due to high memory usage and complex cost volume computations.

\item \textbf{Explicit representation methods}, including Plenoxels\cite{Yu2021PlenoxelsRF}, PlenOctrees\cite{Yu2021PlenOctreesFR}, and DVGO\cite{Sun2021DirectVG}, encapsulate 3D scenes in voxel grids and optimize radiance fields within them. While capable of reproducing intricate geometric and texture details at high resolutions, they cannot generalize to new scenes without extensive per-scene optimization. Methods with learned features \cite{Zhou2018StereoML,Mildenhall2019LocalLF,Srinivasan2019PushingTB,Flynn2019DeepViewVS} achieve better generalization from sparse views but suffer from heavy computational burdens when processing large multi-plane images (MPI) and plane sweep volumes.

\item \textbf{Surrogate geometry methods} improve computational efficiency by sampling around estimated geometry scaffolds, such as depth maps \cite{Deng2021DepthsupervisedNF} and surface meshes. However, they may fail when geometry scaffolds are incorrectly estimated for semi-transparent surfaces or thin structures, limiting their robustness in complex scenarios.
\end{itemize}

These limitations reveal fundamental trade-offs in existing approaches: implicit methods demand high computational resources, explicit methods struggle with generalization, and surrogate geometry methods lack robustness without reliable initial geometric structures.

To address these challenges, we propose \textbf{Hierarchical Sparse Gradient Descent (HSGD)}, which achieves a good trade-off between quality and efficiency in both training and inference stages. And our method leverages the potential geometric and color correspondences in the input plane sweep volumes to achieve generalization, which is free from using some explicit surrogate geometric structures. 

Our key insight is that Multi-plane Images exhibit inherent sparsity in their gradient structure—most gradients have negligible magnitudes and contribute minimally to final rendering quality. By selectively processing only the most significant gradients, HSGD achieves substantial computational acceleration while maintaining visual fidelity. Specifically, The efficiency is achieved by the proposed 2-step processing. Firstly, the multi-plane gradients $\mathcal{A}_d$ are initially obtained by computing the gradients of the MPI with respect to its RGB channel, as outlined in DeepView\cite{Flynn2019DeepViewVS}. Subsequently, $\mathcal{A}_d$ undergoes sparsification by removing small-value gradients. Finally, the sparse gradient descent module updates the MPI based on the sparse gradients.

Experiments on public datasets demonstrate that compared to previous methods\cite{Zhou2018StereoML,Mildenhall2019LocalLF,Flynn2019DeepViewVS}, our optimization strategy based on the sparsity of gradients significantly reduces computational complexity without loss of accuracy. Also, in comparison with methods\cite{Lin2021EfficientNR} relying on surrogate geometry, our method achieves superior visual performance while maintaining high temporal efficiency.

Overall, the technical contributions are summarized as follows.
\begin{itemize}
    \item We present an {optimization strategy: Hierarchical Sparse Gradient Descent (HSGD)}, which efficiently recovers high-quality light fields while effectively reducing inference time and GPU memory usage. 
    \item {Extensive experiments on public datasets demonstrate that \textbf{RealLiFe} achieves a PSNR improvement of approximately \textbf{2dB} compared to other online approaches, or comparable performance while being on average \textbf{100x} faster than offline approaches.}
    \item As a 3D display application demonstrated in Fig. \ref{fig_teaser}, our solution can generate MPIs at above 35 FPS, and novel views can be rendered from a built-up MPI at about 700 FPS for the resolution of $378\times 512$. 
    Novel views of different angles are then provided to a 3D display, which presents 3D effects in the naked eye through multilayer light field decomposition and directional backlighting \cite{Wetzstein2012TensorD}. 
\end{itemize}

\begin{figure*}[!t]
  \centering
  \includegraphics[width=\textwidth]{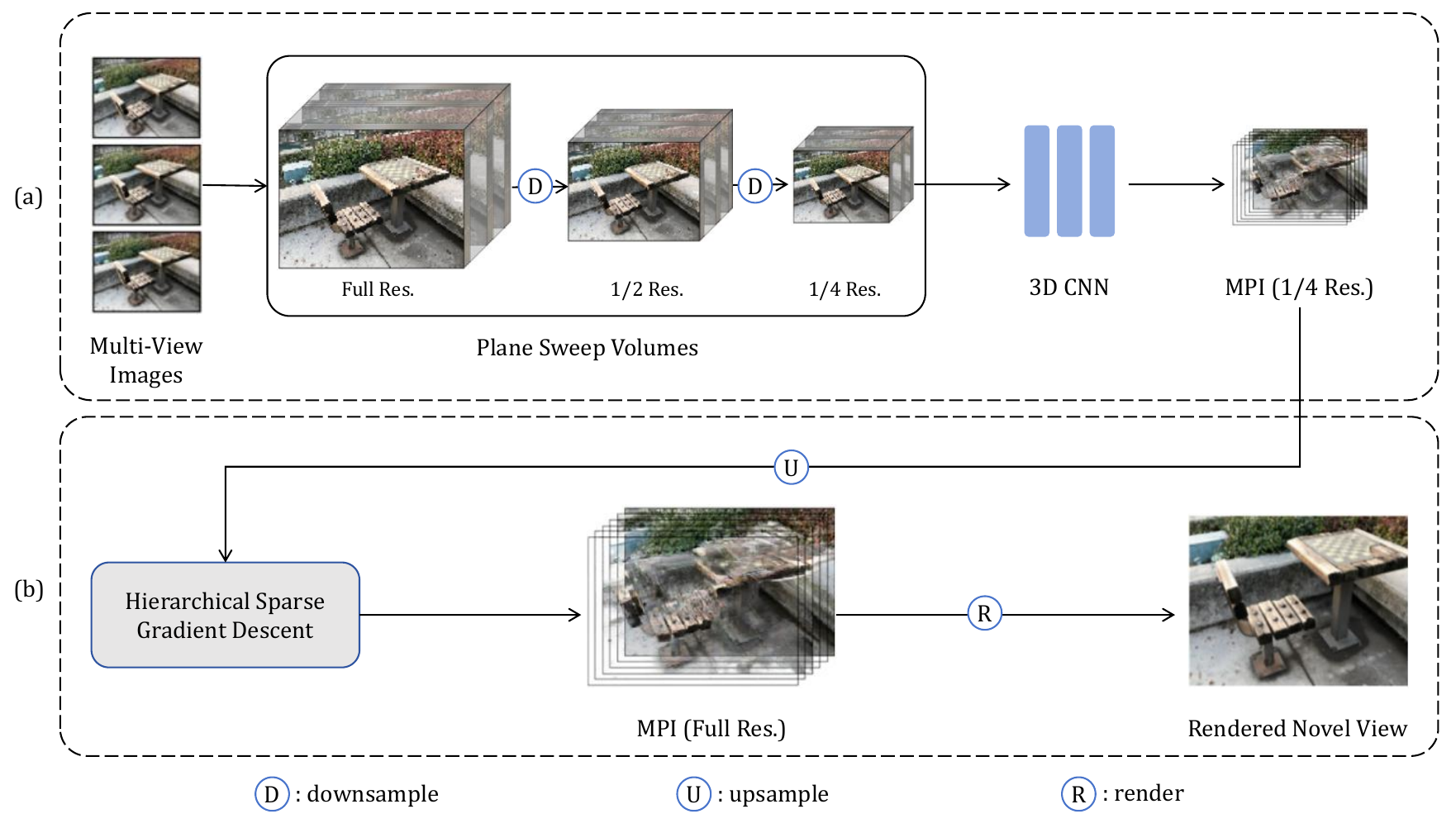}
  \caption{The overview of \textbf{RealLiFe}.(a) Initial MPI Generation: First, the PSV is constructed using multiview images by homographic warping, and the PSV is then downsampled hierarchically at multiple resolutions. The output MPI is then generated in several iterations. Initially, the lowest-resolution PSV is fed to a 3D CNN to extract a coarse-level MPI. (b) Hierarchical Sparse Gradient Descent: The PSV of multi-resolutions and the upsampled MPI both go through the Hierarchical Sparse Gradient Descent module for a refined higher-resolution MPI. Finally, a novel view can be easily rendered from the MPI using the repeated over operator\cite{ThomasKPorter1984CompositingDI}.}
  \label{fig_pipeline}
\end{figure*}
\section{Related Work}
We focus on live light field reconstruction in this paper, which requires novel view synthesis that generalizes to arbitrary scenes with high rendering quality and can be processed in real time. However, existing approaches can hardly achieve all these requirements.

\subsection{High-quality Novel View Synthesis}
The rendering quality of novel view synthesis has seen significant improvement since the proposal of NeRF\cite{Mildenhall2020NeRFRS}. NeRF encodes a scene in a 3D radiance volume and optimizes the density and color of the volume using a simple MLP. Novel views are generated by integrating the volume's density and color along each viewing ray. Based on the success of NeRF, several methods such as Mip-NeRF\cite{Barron2021MipNeRFAM}, RawNeRF\cite{Mildenhall2021NeRFIT}, Ref-NeRF\cite{Verbin2021RefNeRFSV}, and Mip-NeRF360\cite{Barron2021MipNeRF3U} have further improved the rendering quality of the novel view synthesis. For example, Mip-NeRF reduces objectionable aliasing artifacts by casting cones and prefiltering the positional encoding function, while RawNeRF trains the NeRF model on raw camera data to achieve novel high dynamic range view synthesis. Ref-NeRF produces realistic shiny surfaces by explicitly modeling surface reflections, and Mip-NeRF360 solves the problem of unbounded scene novel view synthesis with a non-linear scene parameterization, online distillation, and a novel regularizer. Additionally, another MPI-based method\cite{Zhou2018StereoML} parameterizes each pixel as a linear combination of basis neural functions, successfully reproducing realistic view-dependent effects.

Although these methods have improved the rendering quality from different perspectives, they all require a long per-scene optimization time because they train their models to encode the content of a particular scene, making them not generalizable to new scenes.

\subsection{Generalizable Novel View Synthesis}
To achieve generalizable novel view synthesis, several methods have been proposed that construct radiance fields from image features. PixelNeRF\cite{Yu2021pixelNeRFNR} encodes the input images into pixel-aligned feature grids and then renders points along a target ray by projecting them on the input image. For better performance, IBRNet\cite{Wang2021IBRNetLM} aggregates more information(image colors, features and viewing directions) and designs a ray transformer for decoding colors and densities. GRF\cite{Trevithick2020GRFLA} and NeuRay\cite{Liu2022NeuralRF} achieve occlusion-aware novel view synthesis by incorporating the prior that "The features of a point in 3D space are the same when being viewing from different angles". StereoMag\cite{Zhou2018StereoML}, LLFF\cite{Mildenhall2019LocalLF}, DeepView\cite{Flynn2019DeepViewVS} and MLI\cite{Solovev2022SelfimprovingMI} achieve generalization by modeling the scene as a set of fronto-parallel planes, in which multi-view consistency are subtly encoded. However, MLI\cite{Solovev2022SelfimprovingMI} conducts an extra step to convert the planes to deformable layers and achieves high visual metrics. MVSNeRF\cite{Chen2021MVSNeRFFG} also leverages the concept of multi-view geometry and generalizes to new scenes with a neural cost volume. Instead of introducing image features and geometric priors for generalization, \cite{Tancik2020LearnedIF} applies meta-learning algorithms to learn the initial weight parameters for a NeRF model, enabling faster convergence.

Though generalizable, it still takes them seconds or minutes to generate a novel view. In addition, methods like PixelNeRF\cite{Yu2021pixelNeRFNR} and MVSNeRF\cite{Chen2021MVSNeRFFG} are not robust enough that they usually require extra finetuning to produce satisfying results.

\subsection{Real-time Novel View Synthesis}
To achieve real-time inference speed of a trained scene, structured representations and efficient computation skills are employed. KiloNeRF\cite{Reiser2021KiloNeRFSU} replaces the original large MLP of NeRF with thousands of smaller faster-to-evaluate MLPs, enabling up to 40 FPS of rendering speed.  Plenoctrees\cite{Yu2021PlenOctreesFR} render novel views by representing the scene in a structured octree-based 3D grid. FastNeRF\cite{Garbin2021FastNeRFHN} is capable of rendering novel views at 200 FPS by caching a deep radiance map and efficiently querying it.  Except for NeRF-like representations, a built-up MPI from StereoMag/LLFF/DeepView\cite{Zhou2018StereoML,Mildenhall2019LocalLF,Flynn2019DeepViewVS} can achieve a rendering speed of 60 FPS.  (The rendering speeds of above methods are all evaluated on 800x800 images of Synthetic-NeRF\cite{Mildenhall2020NeRFRS} dataset.)
Despite the real-time inference speed of novel view synthesis, real-time building-up of a new scene is hard to be reached. The cutting-edge optimization-based method: InstantNGP\cite{Mller2022InstantNG} reduces training time of NeRF from hours to several seconds by using a hash grid and hardware acceleration techniques. DeepView\cite{Flynn2019DeepViewVS} is now the fastest MPI-based method, however, it still requires about 50 seconds to generate an MPI. ENeRF\cite{Lin2021EfficientNR} is the state-of-the-art generalizable novel view synthesis method and can render new views at 30 FPS on the ZJU-MoCap\cite{Peng2020NeuralBI} dataset by sampling few points around a computed depth map; however, the rendering speed and image quality can still be improved.

\subsection{Light Field Reconstruction}
Instead of producing a single image at a time, there are methods that output the entire light field of the scene as multi-plane images\cite{Zhou2018StereoML,Mildenhall2019LocalLF,Flynn2019DeepViewVS} or layered mesh representations\cite{Broxton2020ImmersiveLF,Khakhulin2022StereoMW,Solovev2022SelfimprovingMI}. With a generated scene light field, novel views can be rendered easily with nearly no computation budget by simply ray querying. \cite{Broxton2020ImmersiveLF} encodes the scene in a set of multi-sphere images and compresses them into an atlas of scene geometry, allowing light field videos to be streamed over a gigabit connection. Though this method is real-time in rendering and streaming, it still requires a long time to reconstruct a new scene light field. \cite{DeepFusion,SpatialAngular,PixelFeature} optimized for specific neural architectures for improved reconstruction quality of light fields. Their sophisticated designs for the light field capture systems and neural networks result in high metrics in rendering quality but also make them offline light field reconstruction methods. Starline\cite{lawrence2021project} and VirtualCube\cite{Zhang2021VirtualCubeAI} are another two methods that make live light field streaming possible. They both build up their systems using several RGBD cameras, and the additional depth channel enables fast scene geometry extraction, thus saving much time for computation. However, in this paper, we only compare RGB-based novel view synthesis methods to explore their potential for live light field reconstruction.
\section{Method}
The proposed method, \textbf{RealLiFe}, aims to facilitate real-time reconstruction of light fields using a set of sparse posed source view images. To accomplish this goal, we introduce Hierarchical Sparse Gradient Descent (HSGD), an efficient and high-quality optimization technique for MPIs.

\begin{figure*}[!t]
  \centering
  \includegraphics[width=\textwidth]{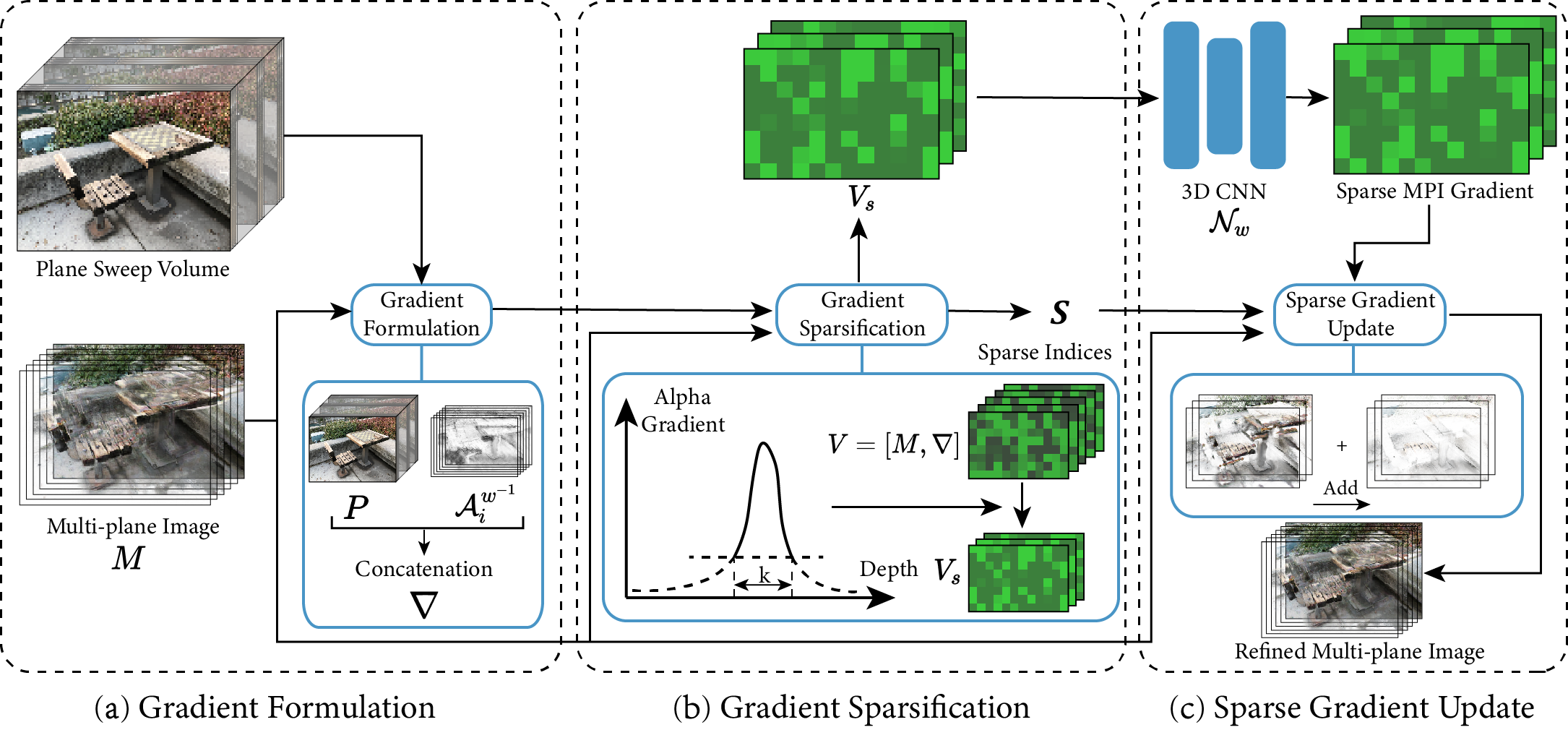}
  \caption{The sparse gradient descent module. 
  (a) The MPI gradients $\nabla$ comprises the input plane sweep volume $P$ and the warped alpha gradients $\mathcal{A}_i^w$ for each source view $i$.
  (b) The volume $V$ is composed of the MPI $M$ and the MPI gradients $\nabla$. It is sparsified by selecting the top $k$ voxels along the depth axis based on alpha gradients of the reference view. Simultaneously, the output sparse indices $S$ store the positions of the selected $k$ voxels along the depth axis.
  (c) The sparsified volume $V_s$ is fed to a 3D CNN (learned gradient descent) for a refiend sparse MPI residual. Finally, the sparse gradient update module utilizes the sparse indices $S$ to add the sparse MPI residual to the input multi-plane image $M$, resulting in a refined multi-plane image.}
  \label{fig_sparse}
\end{figure*}

Fig. \ref{fig_pipeline} presents an overview of the proposed method, which comprises two primary processes: initial MPI generation and hierarchical sparse gradient descent. Sec. \ref{method_dat} introduces the initial MPI generation stage, which encompasses the construction of the plane sweep volume and the initial generation of the MPI. Sec. \ref{method_hier} introduces the hierarchical sparse gradient descent stage, as depicted in Fig. \ref{fig_sparse}. This module comprises three major operations: gradient formulation, gradient sparsification, and sparse gradient update. Prior to discussing our method in detail, we provide a concise introduction of the MPI, and our fundamental optimization method, Learned Gradient Descent, in Sec. \ref{method_pre}. Finally, we provide an introduction of our training approach, with a primary focus on how we construct the loss function, as outlined in Sec. \ref{method_train}.

\subsection{Preliminaries}
\label{method_pre}

\textbf{Multi-plane Image (MPI).} An MPI \cite{Zhou2018StereoML} $M$ is composed of $D$ fronto-parallel $RGB\alpha$ planes in the view frustum of a camera. We denote the RGB channels of an MPI plane as $c_d$ and the alpha channel as $\alpha_d$. The target view $\tilde{I}_t$ is rendered by compositing the MPI in the back-to-front order utilizing the repeated over operator $O$ as defined in \cite{ThomasKPorter1984CompositingDI}:

\begin{equation}
    \tilde{I}_t = \mathcal{O}(\omega_t(M_r)),
    \label{eq_render}
\end{equation}
where $\omega_t$ warps the MPI $M_r$ from reference view to target view via homowarping. And the over operator $\mathcal{O}$ is defined as the compact form:

\begin{equation}
  \mathcal{O}(M) = \sum_{d=1}^{D}c_d\alpha_d\prod_{i=d+1}^{D}(1 - \alpha_i).
  \label{eq_over}
\end{equation}

\textbf{Learned Gradient Descent.} Learned gradient descent is firstly proposed by \cite{Adler2017LearnedPR,Adler2017SolvingII} to solve ill-posed inverse problems for CT reconstruction. Here, we follow DeepView\cite{Flynn2019DeepViewVS} to apply learned gradient descent to the MPI generation problem. Inverse problems refer to problems where one seeks to reconstruct parameters characterizing the system under investigation from indirect observations. Mathematically, the problems can be formulated as reconstructing a signal $f_{true} \in X$ from data $g \in Y$:

\begin{equation}
    g=\mathcal{T}(f_{true})+\delta g,
\end{equation}
where $\mathcal{T}: X\rightarrow Y$ is the forward operator that models how a signal generates data without noise, and $\delta g \in Y$ is a single sample of a $Y$-valued random variable that represents the noise component of data.

Typically, such problems are resolved by iteratively updating the reconstructed $\hat{f}$ with the analytical gradient descent method:

\begin{equation}
    \hat{f}_{n + 1} = \hat{f}_{n} + \lambda [\frac{\partial \mathcal{L}(\hat{f}_{n})}{\partial \hat{f}_{n}} + \frac{\partial \phi (\hat{f}_{n})}{\partial \hat{f}_{n}}],
\end{equation}
where $\mathcal{L}$ is a loss function measuring the difference between the true signal $f_{true}$ and the reconstructed $\hat{f}$, $ \lambda$ is the step size for each iteration, and $\phi$ is a prior on $f_{true}$.

However, it requires too many iterations to converge and is very likely to fall into local optima. Therefore, \cite{Adler2017LearnedPR,Adler2017SolvingII} propose to use learned gradient descent instead of analytical gradient descent, where the update rule is defined by a deep neural network $\mathcal{N}_\omega$:

\begin{equation}
    \hat{f}_{n + 1} = \hat{f}_{n} + \mathcal{N}_\omega [\frac{\partial \mathcal{L}(\hat{f}_{n})}{\partial \hat{f}_{n}} + \hat{f}_{n}].
    \label{eq_learned_theo}
\end{equation}

The learned gradient descent approach is already proved in various works to achieve better results with fewer iterations.

\subsection{Initial MPI Generation}
\label{method_dat}
In this section, we delve into generation of the Initial MPI, a pivotal component in our optimization process, Hierarchical Sparse Gradient Descent. This process involves both the construction of the PSV and a straightforward network forwarding.

\textbf{Hierarchical PSV Construction.} Given $N$ source view images ${I}_{i=1}^N$ of size $H\times W\times 3$ and their corresponding camera poses $[K_i,R_i|t_i]$, as well as for the reference view $I_r$, we construct the plane sweep volume $P$ leveraging inverse homographic warping:
\begin{equation}
    \mathcal{H}^{-1}_i(d) = K_iR_i(\mathcal{I} + \frac{(R_i^{-1}t_i - R_r^{-1}t_r)n^TR_r}{d})R_r^{-1}K_r^{-1},
    \label{eq_homo}
\end{equation}
where $n$ denotes the plane normal and $\mathcal{I}$ is an identity matrix. The inverse homography matrix $\mathcal{H}^{-1}_i(d)$ warps the reference view meshgrid, defined by pixel positions $(u,v)$ at depth $d$, to each source view $I_i$. And the inverse warped source view $I_i^{w^{-1}}$ is grid-sampled by the inverse warped meshgrid:
\begin{equation}
    I_i^{w^{-1}}(u,v,d) = grid\_sample(I_i, \mathcal{H}^{-1}_i(d)[u,v,1]^T),
\end{equation}

By concatenating all the inverse warped source view images $I_i^{w^{-1}}$ along the depth dimension, the plane sweep volume $P$ of size $N \times D \times H \times W \times 3$ is constructed. 
\begin{equation}
    P = [I_0^{w^{-1}}, I_1^{w^{-1}}, ..., I_N^{w^{-1}}]
\end{equation}

To enable optimization at different scales, we downsample it hierarchically at multiple resolutions. Specifically, we create the plane sweep volumes $\{P_l|1<=l<=L\}$ of size $N \times D \times H/2^l \times W/2^l \times 3$ for $L$ iterations. 

\textbf{Network Forwarding.} The coarsest-level PSV is reshaped into $3N \times D \times H/2^l \times W/2^l$ and fed to a 3D neural network to produce the initial MPI. The neural network is trained to aggregate information across views and generate a coarse MPI for subsequent optimization.

\subsection{Hierarchical Sparse Gradient Descent}
\label{method_hier}
Hierarchical sparse gradient descent is a coarse-to-fine approach that progressively refines the MPI using sparse gradients. This design is supported by two key insights: Coarse-to-fine Appearance Refinement and Sparse Gradients Suffice for Light Field Reconstruction. The coarse-to-fine geometry refinement has previously demonstrated its efficacy in various MVS works\cite{Chen2023UncertaintyAW,Jiang2023AdaptMVSNetEM,Weilharter2021HighResMVSNetAF,360MVSNet,Uncertainty,Multistage}. In this context, a low-resolution geometric scaffold can be rapidly derived and employed as a prior to guide the subsequent high-resolution detail refinement process. Similarly, the task of deriving an MPI involves allocating scene contents to their respective depth planes. So the hierarchical pipeline is well-suited for MPI generation, offering a structured foundation upon which the learned gradient descent algorithm can iteratively refine appearance details. As analyzed in~\cite{lin2004geometric}, the light field of a scene can be sufficiently represented by a single dominate plane given accurate depth observations, which indicate that the multi-plane image representation of light field is redundant. 
Hence, any subsequent optimization targeting these redundant regions does not lead to appreciable enhancements in rendering quality but instead incurs unnecessary computational overhead. 


So we propose \textbf{Hierarchical Sparse Gradient Descent}, an adaptive and iterative approach that only leverages sparse gradients for efficient optimization. Specifically, this method consists of three steps: gradient formulation, gradient sparsification, and sparse gradient update. The detailed process is shown in Fig. \ref{fig_sparse}.

\textbf{Gradient Formulation.} To obtain the sparse MPI gradient, we first formulate the components of input gradients. They are defined as the concatenation of the input PSV $P$ and the inverse warped alpha gradient $\mathcal{A}^{w^{-1}}_i$ of each source view $i$. 
Theoretically, the full components of MPI gradients\cite{Flynn2019DeepViewVS} should be in the format of Eq. \ref{eq_origGradients} (The detailed derivation can be referred from \cite{Flynn2019DeepViewVS}.):
\begin{equation}
    \nabla = [P, \widetilde{P}, \mathcal{A}^{w^{-1}}_1,...,\mathcal{A}^{w^{-1}}_N,\mathcal{T}^{w^{-1}}_1,...,\mathcal{T}^{w^{-1}}_N],
    \label{eq_origGradients}
\end{equation}
where $\widetilde{P}$ is the PSV constructed with rendered images at each source view from the generated MPI, and $\mathcal{T}^{w^{-1}}_i$ is the warped MPI gradient respective to MPI's $\alpha$ channel. But we simplify the gradients to be in the format of Eq.~\ref{eq_simplifiedGradients}:
\begin{equation}
    \widetilde{\nabla} = [P, \mathcal{A}^{w^{-1}}_1,...,\mathcal{A}^{w^{-1}}_N],
    \label{eq_simplifiedGradients}
\end{equation}
for several reasons: 

1) The iterative optimization resembles a residual refinement network, where the PSV $P$ is intuitively effective stable and precise residual information directly derived from the input data.

2) The MPI generation process involves the allocation of scene contents to their respective depth planes, with $\mathcal{A}$ being in perfect alignment with the scene contents. This alignment constitutes a crucial gradient component for the accurate construction of MPIs.

3) In the quest for an optimal trade-off between rendering quality and computational efficiency, it is proved in DeepView\cite{Flynn2019DeepViewVS} that $\widetilde{P}$ and $\mathcal{T}$ can only marginally enhance the ultimate rendering quality, but their large spatial sizes can potentially hinder real-time efficiency.

To compute $\mathcal{A}^{w_i^{-1}}$ for a single source view ${i}$, the MPI of the reference view is initially warped to source view $i$ to obtain $M_i$. Then the alpha gradient $\mathcal{A}_{i_d}$ corresponding to depth $d$ is calculated by the following equation: 

\begin{equation}
    \mathcal{A}_{i_d} = \frac{\partial \mathcal{O}(M_i)}{\partial c_d} = \alpha_d\prod_{j=d+1}^{D}(1 - \alpha_j),
    \label{eq_gradient}
\end{equation}
After this, the inverse warped alpha gradient $\mathcal{A}^{w_i^{-1}}$ is derived by applying inverse homographic warping Eq. \ref{eq_homo} from the source view to the reference view. 

After formulating the gradients, the full 3D volume $V$ to be sparsified and optimized is the concatenation of $\widetilde{\nabla}$ and $M$:

\begin{equation}
    V = [\widetilde{\nabla}, M],
    \label{eq_V}
\end{equation}

\label{method_sparsify}
\textbf{Gradient Sparsification.}
To safely sparsify gradient components without adversely affecting the final rendering quality, it becomes imperative to establish criteria for the exclusion of voxels that impart minimal contributions. A commonly adopted strategy in various Multiview Stereo (MVS) studies\cite{Chen2023UncertaintyAW,Jiang2023AdaptMVSNetEM,Weilharter2021HighResMVSNetAF} is to reduce the number of depth layers in the full cost volume from $D$ to $k$, where $k$ is usually an odd number, composed of the plane with the highest probability and its adjacent $2\times(k - 1)/2$ layers. However, this \textit{MVS-sampling} strategy primarily relies on surface-centric geometric considerations, which might be too aggressive to capture some light field details like semi-transparent objects and thin plates. If we revisit Eq. \ref{eq_over}, it can be rewritten as:
\begin{equation}
  \mathcal{O}(M) = \sum_{d=1}^{D}c_d\mathcal{A}_d,
  \label{eq_overA}
\end{equation}
from which we can easily find that the influence of the color component $c_d$ of each MPI plane on the ultimate color is determined by the alpha gradient derived from the reference view $\mathcal{A}_d$. The alpha gradient aligns more closely with the scene contents, as it takes into consideration not only solid geometry but also semi-transparent surfaces and thin plates, making it a more comprehensive and effective choice. In the ablation study of Sec. \ref{sec_ablation}, we demonstrate that our sparsification strategy consistently outperforms the \textit{MVS-sampling} approach in terms of overall performance. Consequently, the criterion for gradient sparsification is determined by $\mathcal{A}_d$. 

\begin{figure*}[!t]
    \centering
    \includegraphics[width=\textwidth]{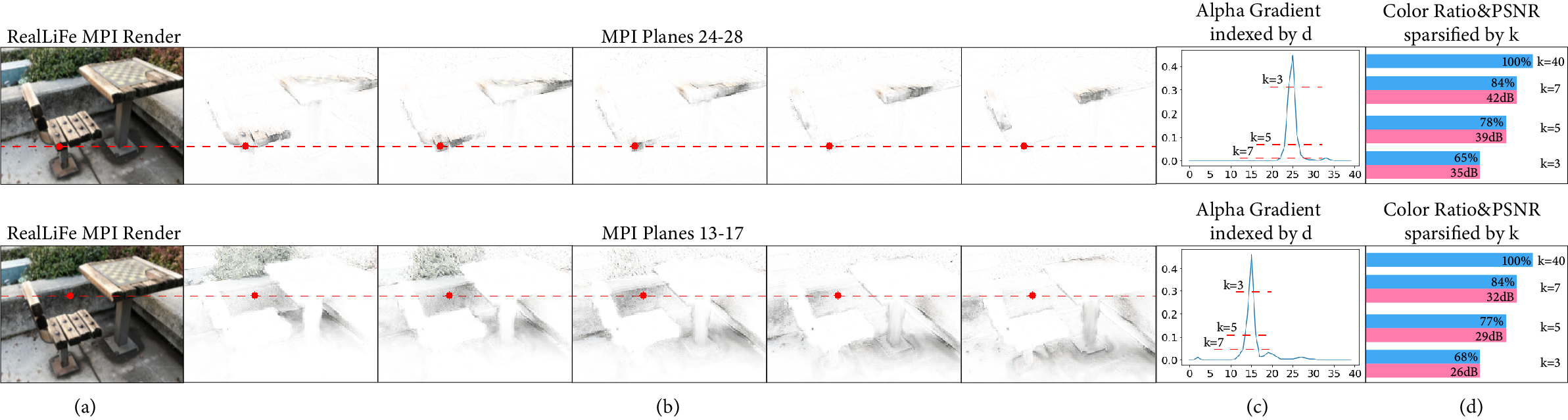}
    \caption{The influence of $k$ on the rendering quality. (a) A rendered image of \textbf{RealLiFe} (without using HSGD) with 40 MPI planes. (b) Top 5 MPI planes with the highest alpha gradients $\mathcal{A}$ for the red pixel. (The MPI is multiplied by $\mathcal{A}$ to better visualize its contribution to the final rendering result.) (c) The alpha gradients of the red pixel across 40 MPI planes, with the red dotted lines partitioning $d$ based on whether its alpha gradient falls within the top $k$. (d) The color ratio ($RGB_{k=3,5,7} / RGB_{k=40}$) and PSNR of the rendered red pixel in comparison to when $k=40$.}
    \label{fig_k}
\end{figure*}

The remaining challenge is to decide the optimal $k$ for gradient sparsification. Fig. \ref{fig_k} provides insights into the determination of the optimal value for $k$. For instance, when $k=7$, the rendered color almost recovers more than $80\%$ of the original color, indicating a trade-off between computational efficiency and rendering quality. Generally, it appears from Fig. \ref{fig_k} that values of $k$ in the range of 5 to 7 represent favorable choices, achieving a balance between time efficiency and rendering quality. Further ablation experiments on different values of $k$ will be presented in Sec. \ref{sec_ablation}. 

After sparsification, only $k$ voxels with the highest alpha gradient along the depth axis are kept, resulting in the compression of the input volume $V$, which is the concatenation of the original gradient volume $\widetilde{\nabla}$ and the input MPI $M$, from $C\times D\times H\times W$ to $C\times k\times H\times W$. 
The sparse indices $S$, which store the positions of the $k$ selected voxels in the original depth axis, are also retained to enable the generated spasified MPI residual to be added to the input complete MPI for an update. It is evident that while the complete spatial structure of the light field is sacrificed after sparsification, the sparsified layers are retained in their original spatial order, preserving the potential surface structure of the light field for subsequent optimization.

\label{method_update}
\textbf{Sparse Gradient Update.} After the gradient components of each voxel has been decided, a network backbone needs to be decided for the learned gradient descent module. A CNN with large perceptive fields is proved effective in DeepView\cite{Flynn2019DeepViewVS}. However, after gradient sparsification, an issue arises where adjacent voxels may not necessarily belong to the same depth plane. This raises concerns about the suitability of convolutional operations on these voxels. So a straightforward decision for the backbone would be an MLP, which treats each voxel independently. However, this choice compromises the benefits of an extensive receptive field, resulting in suboptimal performance. Moreover, even though the depths of a selected subset of voxels may exhibit considerable disparities, convolutional processing enables the model to adapt to these variations by learning shared features and patterns across different regions. We further prove the adaptability of the 3D CNN in enhancing rendering quality without introducing conspicuous artifacts through qualitative analysis in Sec. \ref{sec_qualitative} and quantitative evaluation in Sec. \ref{sec_quantitative}.

As shown in Fig. \ref{fig_sparse} (c), the sparsified input volume $V_s$ goes through a 3D CNN, functioning as the learned gradient descent network, to derive the sparse MPI residual for an MPI refinement iteration.
The voxels in the sparse MPI residual are then restored to their original positions using the sparse indices $S$, and subsequently added to the input MPI to obtain a refined version:
\begin{equation}
    M_{n+1} = M_n + \mathcal{R}(\mathcal{N}_w(V_s), S),
\end{equation}
where $\mathcal{N}_w$ is the 3D CNN that functions as the learend gradient descent neural network, and $\mathcal{R}$ is the operator that restored the voxels of the sparse MPI residual to their original positions.

\subsection{Training}
\label{method_train}
During training, the network parameters are progressively optimized by minimizing the difference between the synthesized views and the ground truth novel views. And we choose the commonly used deep feature matching loss $L_{VGG}$\cite{QifengChen2017PhotographicIS} as the basic loss function. In detail, our overall rendering loss $L_{r}$ is a weighted average loss of all iterations, and the rendering loss for each iteration is also a weighted average one for both the synthesized reference and source views:
\begin{equation}
    L_r = \sum_{i=1}^{l}\lambda_i(L_{VGG}(I_{r_i}, \tilde{I}_{r_i}) + 
    \mu\sum_{j=1}^{N}L_{VGG}(I_{j_i}, \tilde{I_{j_i}})),
\end{equation}
where $I_{r_i}$ and $I_{j_i}$ are the ground truth reference 
and source images at iteration $i$, $\tilde{I}_{r_i}$ and $\tilde{I}_{j_i}$ are the rendered reference and source images from the generated MPI at iteration $i$. And $l$ is the number of iterations, $N$ is the number of input source images, $\lambda_i$ is a hyper-parameter weighing the importance of each iteration, and $\mu$ is also a hyper-parameter balancing reference and source image supervision. 

In addition, we propose a sparsity loss $L_s$ that regularizes the alpha values of the MPI to be close to 0 or 1:
\begin{equation}
    L_s = \frac{\sum_{h=1}^{H}\sum_{w=1}^{W}\sum_{d=1}^{D}log(1.5 - |0.5 - \alpha_{h,w,d}|)}{H\times W\times D},
\end{equation}
where $\alpha_{h,w,d}$ is the alpha channel of the MPI, and $H, W, D$ are the height, width and number of depth planes of the volume. This loss aims to reduce the entropy of the MPI with respect to its $\alpha$ channel. It is ideal for gradient sparsification, as it allows us to select more informative top $k$ samples that lie around the surfaces of a scene. The overall training loss is:
\begin{equation}
    L = L_r + \lambda_s L_s,
\end{equation}
where $\lambda_s$ is a hyper-parameter balancing color supervision and sparsity regularization.

\section{Implementation Details}
\begin{figure}[!t]
  \centering
  \includegraphics[width=\columnwidth]{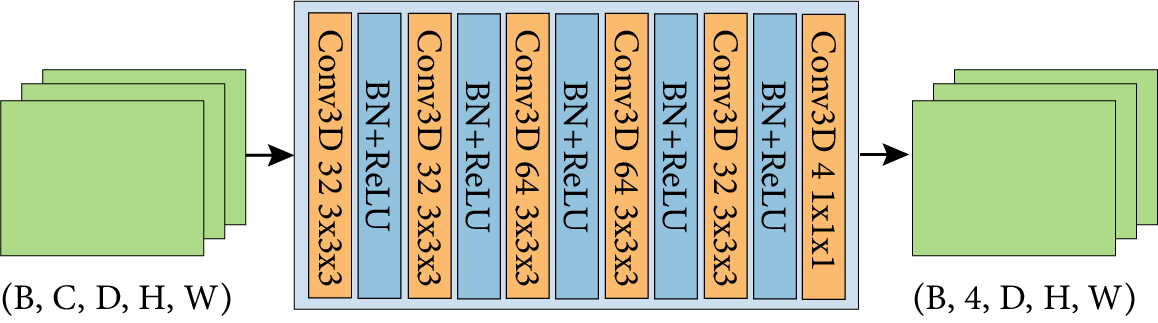}
  \caption{The backbone 3D CNN of \textbf{RealLiFe}, which contains only six convolution layers.}
  \label{fig_net}
\end{figure}

The generalized model was trained using an RTX 3090 GPU with the Adam optimizer\cite{Kingma2014AdamAM}. The initial learning rate was configured at $1e-3$, with a reduction by half if the training loss consistently decreased for a continuous span of 10 epochs. The model underwent training for 1000k iterations, taking approximately 13 hours to complete. The trainable components within the entire pipeline included the 3D CNNs responsible for converting the input plane sweep volume into a multi-plane image (as depicted in Fig. \ref{fig_pipeline}) and refining the sparse MPI gradient (as illustrated in Fig. \ref{fig_sparse}). The structure of the 3D CNN, as shown in Fig. \ref{fig_net}, comprised 5 basic convolutional blocks (a convolution layer followed by a batchnorm layer and a relu layer) and 1 1x1x1 3D convolution layer. To further improve the efficiency of our method, we optimize to run our pipeline with TensorRT\footnote{https://developer.nvidia.com/tensorrt}.

To configure the loss function hyperparameters, a preference was given to increased color supervision from the reference view and the last iteration, while maintaining the impact of the sparsity loss. An example set of these parameters was as follows: $\lambda_1=0.2$, $\lambda_2=0.3$, $\lambda_3=0.5$, $\mu=0.5$, $\lambda_s=0.2$. Regarding experimental parameters, the model utilized 40 MPI planes, underwent 3 network iterations, and employed a sparsification factor $k$ set to 5 as the default. These settings may vary in the context of ablation experiments.


\begin{table*}[t]
\caption{Quantitative comparison on SWORD\cite{Khakhulin2022StereoMW} (25 scenes), Real Forward-Facing\cite{Mildenhall2019LocalLF} (8 scenes) and Shiny\cite{wizadwongsa2021nex} (8 scenes) evaluation dataset. The rank-1 method's metrics are highlighted in deep green, while those of the rank-2 method are highlighted in light green. DeepView (Re-Imp) denotes the re-implemented version of DeepView.}
\centering
\resizebox{\textwidth}{!}{
\begin{tabular}{*{2}{l}*{14}{c}} %
    \toprule
    \multirowcell{2}[0pt][l]{Category} & \multirowcell{2}[0pt][l]{Model} & \multirowcell{2}[0pt][c]{Generation \\Time, sec} & \multirowcell{2}[0pt][c]{Offline \\Rendering, fps} & \multirowcell{2}[0pt][c]{3D Display \\Efficiency, fps} & \multicolumn{3}{c}{\textbf{SWORD} } && \multicolumn{3}{c}{\textbf{Real Forward-Facing} } && \multicolumn{3}{c}{\textbf{Shiny} } \\
    \cmidrule{6-8} \cmidrule{10-12} \cmidrule{14-16} 
    &&&&& PSNR $\uparrow$ & SSIM $\uparrow$ & LPIPS $\downarrow$  && PSNR $\uparrow$ & SSIM $\uparrow$ & LPIPS $\downarrow$ && PSNR $\uparrow$ & SSIM $\uparrow$ & LPIPS $\downarrow$ \\
    \midrule
    \multirowcell{3}{Offline}
    & IBRNet\cite{Wang2021IBRNetLM} & 4.174 & - & $\sim$0.01 & 23.04 & 0.77 & 0.23 && 23.64 & 0.78 & 0.21 && 24.20 & 0.82 & 0.17 \\
    & MVSNeRF\cite{Chen2021MVSNeRFFG} & 1.523 & - & $\sim$0.04 & 21.79 & 0.74 & 0.24 && 22.25 & 0.75 & 0.21 && 23.33 & 0.83 & 0.14 \\
    & SIMPLI-8L\cite{Solovev2022SelfimprovingMI} & 1.677 & \cellcolor[HTML]{addb88}$\sim$480.0 & $\sim$0.60 & \cellcolor[HTML]{369f2d}24.22 & \cellcolor[HTML]{369f2d}0.83 & \cellcolor[HTML]{369f2d}0.13 && 25.01 & \cellcolor[HTML]{369f2d}0.86 & \cellcolor[HTML]{369f2d}0.09 && \cellcolor[HTML]{369f2d}27.27 & \cellcolor[HTML]{369f2d}0.91 & \cellcolor[HTML]{369f2d}0.06 \\
    & DeepView\cite{Flynn2019DeepViewVS} (Re-Imp) & 10.081 & $\sim$700.0 & $\sim$0.10 & 21.99 & 0.73 & 0.18 && 23.11 & 0.76 & 0.13 && 24.99 & 0.81 & 0.09 \\
    \midrule
    \multirowcell{3}{Online}
    & ENeRF\cite{Lin2021EfficientNR} & 0.030 & - & $\sim$1.85 & 22.95 & 0.75 & 0.19 && 22.90 & 0.77 & 0.18 && 26.03 & 0.86 & 0.10 \\
    & \textbf{RealLiFe} & \cellcolor[HTML]{addb88}0.026 & $\sim$700.0 & \cellcolor[HTML]{addb88}$\sim$37.04 & \cellcolor[HTML]{addb88}23.82 & \cellcolor[HTML]{addb88}0.77 & \cellcolor[HTML]{addb88}0.20 && \cellcolor[HTML]{369f2d}25.46 & \cellcolor[HTML]{addb88}0.84 & \cellcolor[HTML]{addb88}0.15 && \cellcolor[HTML]{addb88}27.15 & \cellcolor[HTML]{addb88}0.85 & \cellcolor[HTML]{addb88}0.10 \\
    & \textbf{RealLiFe-2I} & \cellcolor[HTML]{369f2d}0.022 & \cellcolor[HTML]{369f2d}$\sim$700.0 & \cellcolor[HTML]{369f2d}$\sim$45.45 & 23.80 & 0.77 & 0.20 && \cellcolor[HTML]{addb88}25.33 & 0.84 & 0.16 && 27.03 & 0.85 & 0.10 \\
    \bottomrule
    \end{tabular}
}
\vspace{0.7ex}
\label{tab_main}
\end{table*}


\begin{table}[htbp]
    \caption{Quantitative comparison on Spaces evaluation dataset in terms of SSIM. The rank-1 method's metrics are highlighted in deep green, while those of the rank-2 method are highlighted in light green.}
    \centering
    \resizebox{\columnwidth}{!}{
    \begin{tabular}{*{1}{l}*{6}{c}}
        \toprule
        Configuration & Soft3D\cite{Penner2017Soft3R} & StereoMag$^{+}$\cite{Zhou2018StereoML} & DeepView\cite{Flynn2019DeepViewVS} & \textbf{RealLiFe} & \textbf{RealLiFe-2I} \\
        \midrule
        Small baseline & 0.9260 & 0.8884 & \cellcolor[HTML]{369f2d}0.9541 & \cellcolor[HTML]{addb88}0.9396 & 0.9286 \\
        Medium baseline & 0.9300 & 0.8874 & \cellcolor[HTML]{369f2d}0.9544 & \cellcolor[HTML]{addb88}0.9317 & 0.9159 \\
        Large baseline & \cellcolor[HTML]{addb88}0.9312 & 0.8673 & \cellcolor[HTML]{369f2d}0.9485 & 0.9180 & 0.9052 \\
        \midrule
        Generation Time, sec & - & - & $\sim$20.00 & \cellcolor[HTML]{addb88}0.044 & \cellcolor[HTML]{369f2d}0.037 \\
        \bottomrule
    \end{tabular}
    }
    \label{tab_spaces}
\end{table}

\section{Experiments}
\subsection{Baselines, datasets and metrics}

\textbf{Baselines.} In order to evaluate the visual performance and real-time efficiency of our method, we compare it with four generalizable novel view synthesis methods: IBRNet\cite{Wang2021IBRNetLM}, MVSNeRF\cite{Chen2021MVSNeRFFG}, and ENeRF\cite{Lin2021EfficientNR}. Additionally, we compare our method with four light field reconstruction methods, namely Stereo Magnification (StereoMag)\cite{Zhou2018StereoML}, Soft3D\cite{Penner2017Soft3R}, MLI\cite{Solovev2022SelfimprovingMI}, and DeepView\cite{Flynn2019DeepViewVS}. Among the baseline methods, IBRNet, MVSNeRF and ENeRF conduct extra finetuning experiments to improve performance. However, finetuning takes so much time that it is incompatible with real-time light field reconstruction, thus we only compare the results without finetuning. LLFF\cite{Mildenhall2019LocalLF} is excluded as a baseline method because the number of input views for the released model is set fixed to 5. Additionally, LLFF fuses multiple MPIs to render novel views, leveraging information of up to 10-20 input views, making it a method with dense view inputs. However, it is essential to acknowledge that LLFF\cite{Mildenhall2019LocalLF} remains a robust approach for generating high-quality light fields offline.

\textbf{Datasets.} For training and evaluation, we select 4 datasets: Spaces\cite{Flynn2019DeepViewVS}, Real Forward-Facing\cite{Mildenhall2019LocalLF}, SWORD\cite{Khakhulin2022StereoMW} and Shiny\cite{wizadwongsa2021nex}. To compare results on the Real Forward-Facing evaluation set, comprising 8 scenes, the SWORD evaluation set, comprising 25 scenes (selected from the total evaluation scenes) and Shiny evaluation set, comprising 8 scenes, we trained our method using a combined training set in the same manner as IBRNet\cite{Wang2021IBRNetLM}. This combined training set consists of 90 scenes from the Spaces training set and 35 scenes from the Real Forward-Facing training set. To compare results on the Spaces evaluation set, which consists of 10 scenes with three settings of large, medium, and small baselines, we trained our method using the same training dataset as DeepView\cite{Flynn2019DeepViewVS}, comprising 90 scenes from the Spaces training set.

\textbf{Metrics.} We compared the visual performance of the proposed method using standard metrics, structural similarity (SSIM), peak signal-to-noise ratio (PSNR), and perceptual similarity (LPIPs). We evaluated the temporal efficiency of all methods based on \textit{Generation Time}, the time required to generate a single representation (RGB images for IBRNet\cite{Wang2021IBRNetLM}, MVSNeRF\cite{Chen2021MVSNeRFFG} and ENeRF\cite{Lin2021EfficientNR}, MPIs for Soft3D\cite{Penner2017Soft3R}, StereoMag\cite{Zhou2018StereoML}, DeepView\cite{Flynn2019DeepViewVS} and our method, multi-layered meshes for MLI\cite{Solovev2022SelfimprovingMI}). Furthermore, given that light field reconstruction methods yield light field representations suitable for offline rendering, we compared the offline rendering speed of light field reconstruction methods, also in FPS. (We use the OpenGL renderer provided by LLFF\cite{Mildenhall2019LocalLF} to evaluate the offline rendering speed.) To assess the viability of each method for 3D display applications, we evaluated their \textit{3D display efficiency} (measured in FPS), which we define as the reciprocal of the time required to provide sufficient light field information to support light field rendering on a 3D display. Specifically, a 3D display (e.g., a looking glass) requires $n$ views ($n$ is set to 18 in Tab. \ref{tab_main}) at different viewing angles to reproduce a light field, providing a continuous, forward-facing viewing experience. Novel view synthesis methods necessitate $n$ forward processes to generate $n$ new views. In contrast, light field reconstruction methods require only a single forward process to generate $n$ novel views because the built-up representations enable fast novel view rendering in the forward-facing viewing range.

\subsection{Experiment Configurations}
We compare our method with state-of-the-art generalizable novel view synthesis methods and light field reconstruction methods on public datasets including SWORD\cite{Khakhulin2022StereoMW}, Real Forward-Facing\cite{Mildenhall2019LocalLF}, Shiny\cite{wizadwongsa2021nex} and Spaces \cite{Flynn2019DeepViewVS}. 

\textbf{SWORD, Real Forward-Facing and Shiny.} To compare results on a sparse view setting, we set the number of input views to 3 for all methods. The evaluation image resolution for three datasets is $378\times 512$, and the number of MPI planes for our method is set to 40. To compare with MLI\cite{Solovev2022SelfimprovingMI}, we choose their best-performance model SIMPLI-8L that uses 8 layered meshes for representations. We trained two versions of our models, RealLiFe and RealLiFe-2I. RealLiFe has 3 iterations of learned gradient descent, while RealLiFe-2I has 2 iterations, directly upsamping the 1/4-scale MPI to the original scale. In order to assess the interpolation and extrapolation capabilities, we evaluate all views from the three datasets. For each view of a given scene, we select the nearest available 3 views as input. The evaluation results for a single scene are calculated by averaging the results across all views. Likewise, the final evaluation results for a dataset are computed by averaging the results across all scenes.

\textbf{Spaces.} We follow the configuration of DeepView\cite{Flynn2019DeepViewVS} to compare with their evaluation results. The number of input views is 4, the evaluation image resolution is $480\times 800$, and the number of planes of an MPI is 40. We use the same input and evaluation views as DeepView\cite{Flynn2019DeepViewVS} for three baseline settings. 

Our models are trained for 100,000 iterations for both configurations, and all methods are evaluated on an RTX 3090 GPU.

\label{sec_quantitative}
\subsection{Quantitative Results}
Tab. \ref{tab_main} presents the quantitative results on SWORD\cite{Khakhulin2022StereoMW}, Real Forward-Facing\cite{Mildenhall2019LocalLF} and Shiny\cite{wizadwongsa2021nex}. We classify all methods into offline and online according to \textit{Generation time}. 

Compared with offline methods, our model RealLiFe and RealLiFe-2I stands out in terms of visual metrics, achieving top-1 or top-2 rankings on three datasets. And our default model RealLiFe strikes a balance between the rendering quality and efficiency. Our models, on average, achieve a $100\times$ faster generation speed compared to other offline methods. Compared with the online method. Our models are a little bit faster than ENeRF\cite{Lin2021EfficientNR} in generation time; however, our method exhibits superior visual metrics on three datasets. When considering the metric of 3D Display Efficiency, our method stands out as the only one capable of supporting real-time light field video display. This is due to its high temporal efficiency in both light field reconstruction and offline rendering speed.

Tab. \ref{tab_spaces} lists the quantitative results on the Spaces evaluation set, where we can see that DeepView generates the highest-quality rendering results overall, followed by our method and Soft3D\cite{Penner2017Soft3R}. The decrease in rendering quality may be attributed to the fact that our method depends on network connections to propagate visibility information. As we increase the baseline, there is a corresponding increase in the number of network connections required to effectively propagate visibility, as noted in \cite{Flynn2019DeepViewVS}. We have also reported the results of our re-implemented version of DeepView, however, the results show some gaps with the original DeepView, this is mainly due to the visual discontinuities between small MPI patches. Since DeepView require high GPU memory to infer a full-sized MPI, so we have to infer small MPI patch by patch and then merge them all into a complete one, and this process leads to performance drop in our re-implemented version. However, in terms of time efficiency, our method is about 400 times faster than DeepView. The \textit{Generation Time} of Soft3D and StereoMag$^{+}$ are left as "-" because Soft3D is not open-sourced and StereoMag$^{+}$ is a modified version by the authors of DeepView, so that we can not evaluate their time efficiency accurately. To summarize, we have the following observations:
\begin{itemize}
    \item Compared with offline novel view synthesis methods\cite{Wang2021IBRNetLM,Chen2021MVSNeRFFG} and offline light field reconstruction methods\cite{Zhou2018StereoML, Penner2017Soft3R,Flynn2019DeepViewVS,Solovev2022SelfimprovingMI}, our approach achieves around $100\times$ speedup ratio with superior or comparable visual performance.
    \item Compared with the online novel view synthesis method\cite{Lin2021EfficientNR}, our method achieves better visual performance, while maintaining comparable time efficiency.
    \item Novel views can be rendered at significantly higher FPS than other novel view synthesis methods\cite{Wang2021IBRNetLM,Chen2021MVSNeRFFG,Lin2021EfficientNR} from our generated MPIs. This property makes our method particularly suitable for displaying on 3D displays and other light mobile devices.
\end{itemize}

\begin{figure*}
    \centering
    \includegraphics[width=\textwidth]{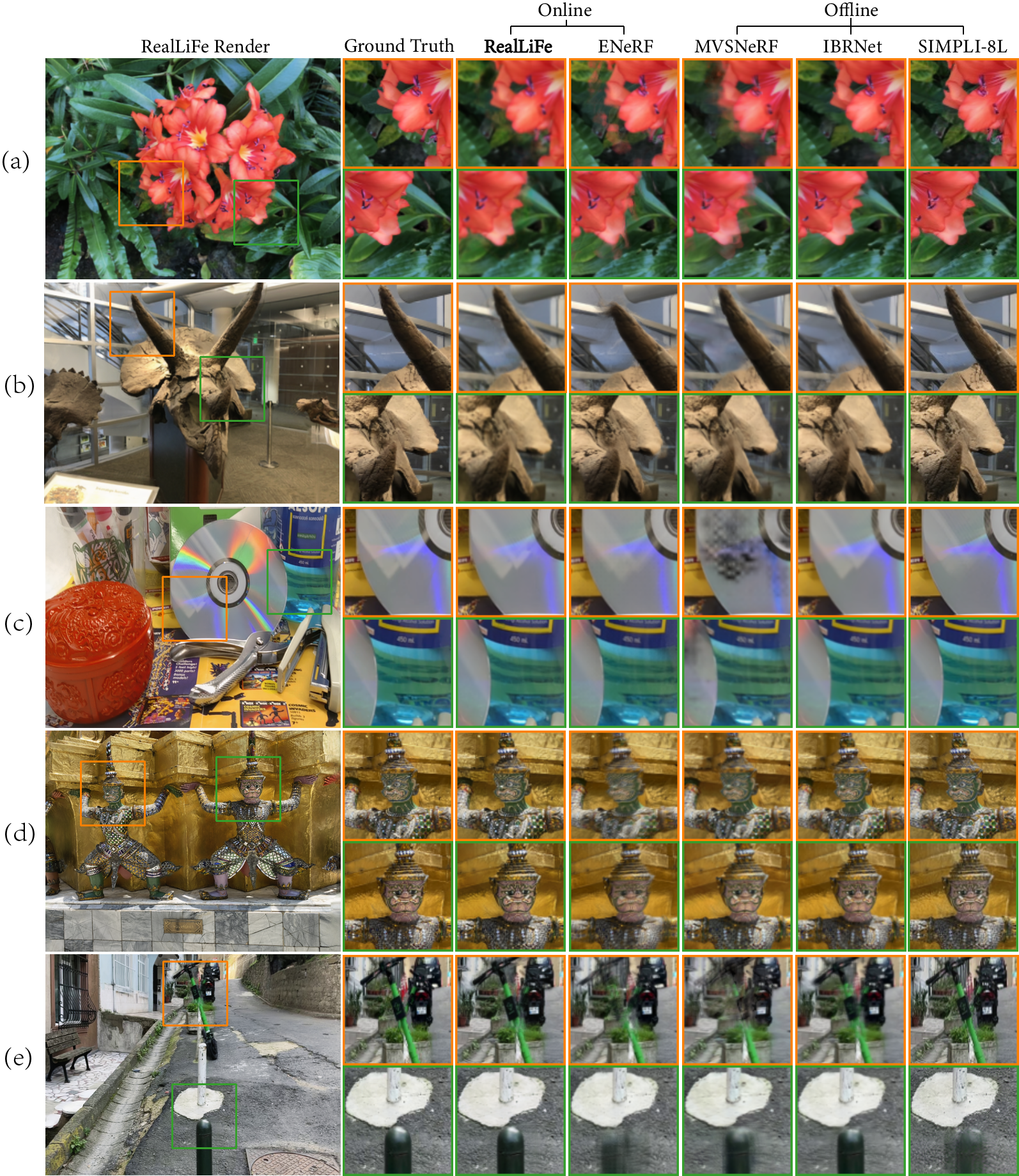}
    \caption{Qualitative results on Real Forward-Facing\cite{Mildenhall2019LocalLF} (a) and (b), Shiny\cite{wizadwongsa2021nex} (c) and (d) and SWORD\cite{Khakhulin2022StereoMW} (e) evaluation datasets. }
    \label{fig_exp_llff}
\end{figure*}

\begin{figure*}
    \centering
    \includegraphics[width=\textwidth]{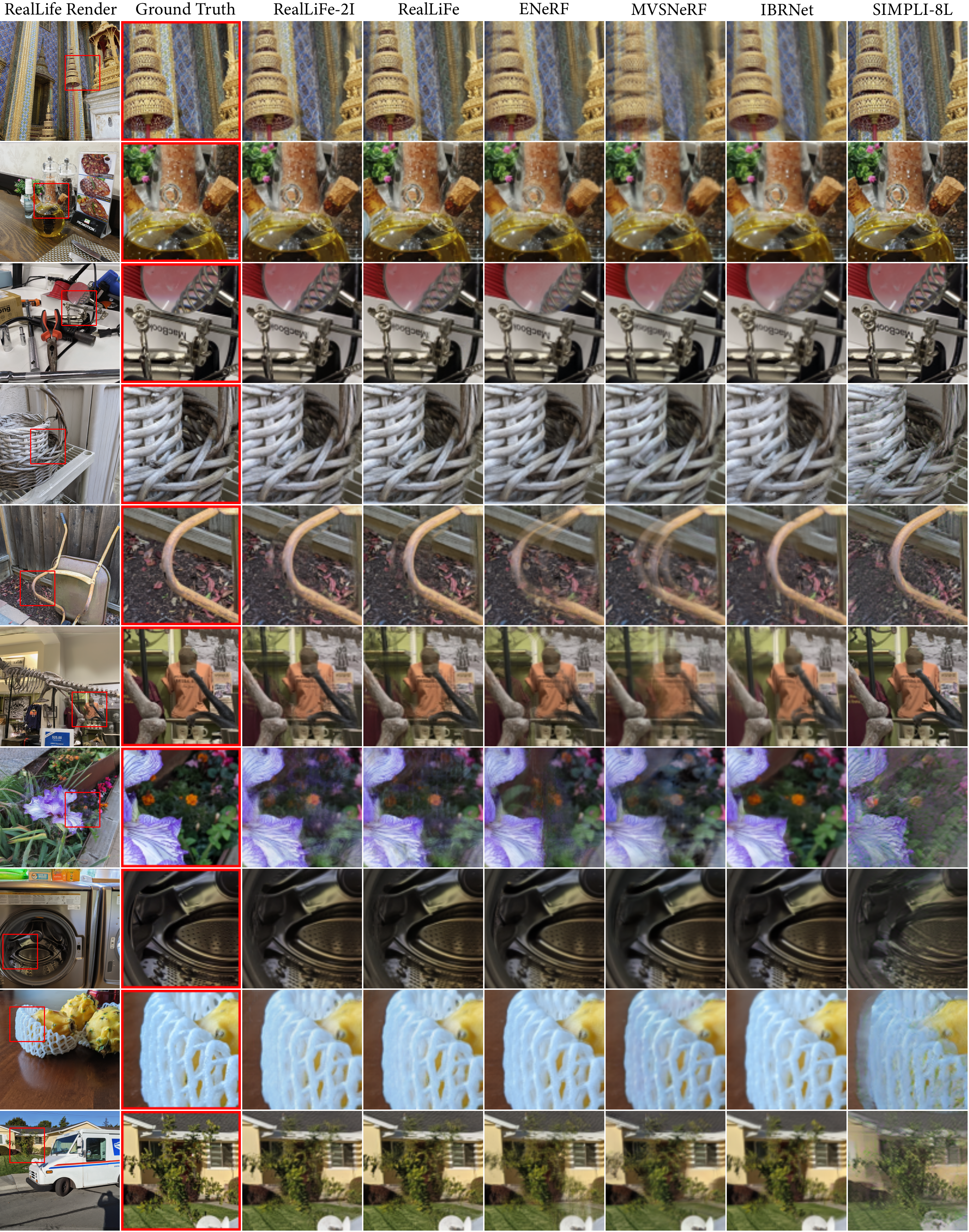}
    \caption{Extra qualitative comparison on Shiny\cite{Solovev2022SelfimprovingMI} and IBRNet collected\cite{Wang2021IBRNetLM}. }
    \label{fig_extra1}
\end{figure*}

\begin{figure*}
    \centering
    \includegraphics[width=\textwidth]{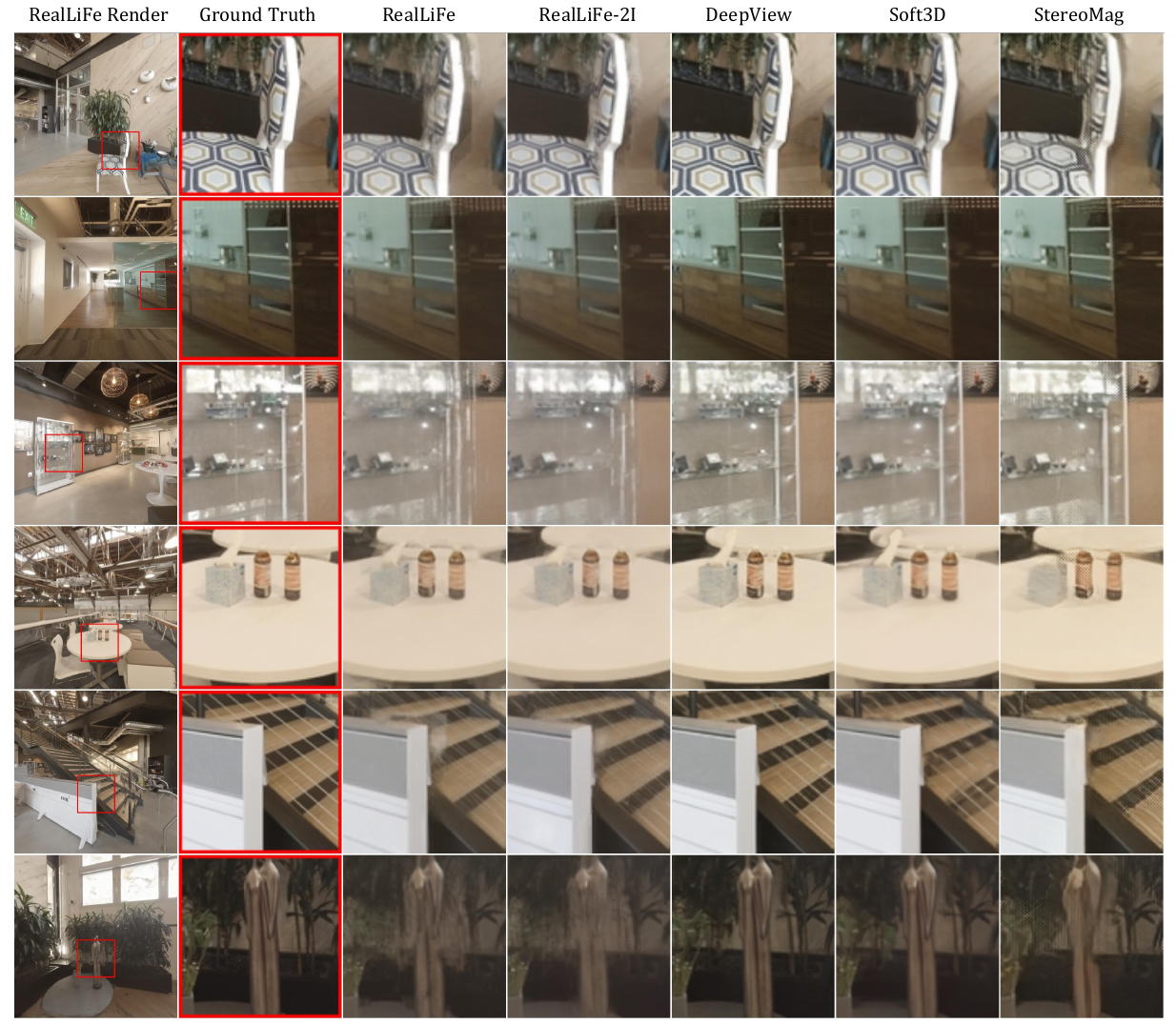}
    \caption{Qualitative comparison on Spaces\cite{Flynn2019DeepViewVS} dataset.}
    \label{fig_spaces}
\end{figure*}

\label{sec_qualitative}
\subsection{Qualitative Results}
We show qualitative comparisons on Real Forward-facing\cite{Mildenhall2019LocalLF} (the first two rows), Shiny\cite{wizadwongsa2021nex} (the third row) and SWORD\cite{Khakhulin2022StereoMW} (the last row) in Fig. \ref{fig_exp_llff}. 
For a comprehensive comparison with previous state-of-the-art methods, we evaluate the rendering quality in the following aspects.

\textbf{Occluded Regions.} In Fig. \ref{fig_exp_llff} (a), ghosting effects are produced at the edges of flowers by ENeRF\cite{Lin2021EfficientNR} and MVSNeRF\cite{Chen2021MVSNeRFFG}. In contrast, our method generates clear boundaries, comparable to IBRNet\cite{Wang2021IBRNetLM} and SIMPLI-8L\cite{Solovev2022SelfimprovingMI}. In Fig. \ref{fig_exp_llff} (b), ENeRF\cite{Lin2021EfficientNR} produces an incomplete horn, and MVSNeRF\cite{Chen2021MVSNeRFFG} blurs the background around the horn. Our results are comparable to IBRNet\cite{Wang2021IBRNetLM}, which show a complete horn and clean background, but not as good as SIMPLI-8L\cite{Solovev2022SelfimprovingMI}.

\textbf{Intricate Texture Details.} Fig. \ref{fig_exp_llff} (d) shows the faces of two statues with intricate texture details. ENeRF\cite{Lin2021EfficientNR}, MVSNeRF\cite{Yao2018MVSNetDI}, and IBRNet\cite{Wang2021IBRNetLM} produce varying degrees of blurriness, while our method and SIMPLI-8L\cite{Solovev2022SelfimprovingMI} restore the complicated texture details very well.

\textbf{View-dependent Effects.} Fig. \ref{fig_exp_llff} (c) and (e) show view-dependent specularity effects on the cd and the black pillar, respectively. For the cd scene, MVSNeRF\cite{Chen2021MVSNeRFFG} fails to produce reasonable specular patterns, and ENeRF\cite{Lin2021EfficientNR} and SIMPLI-8L\cite{Solovev2022SelfimprovingMI} generate blurry edges of the white reflection. However, our method and IBRNet\cite{Wang2021IBRNetLM} produce more accurate view-dependent effects. For the black pillar, ENeRF\cite{Lin2021EfficientNR}, SIMPLI-8L\cite{Solovev2022SelfimprovingMI}, and MVSNeRF\cite{Chen2021MVSNeRFFG} fail to generate the complete shape of the pillar, and IBRNet\cite{Wang2021IBRNetLM} does not restore the specularity as accurately as our method.

\textbf{Thin Plates.} In Fig. \ref{fig_exp_llff} (e), the green stick of the scooter is a very thin plate. ENeRF\cite{Lin2021EfficientNR}, MVSNeRF\cite{Chen2021MVSNeRFFG}, and IBRNet\cite{Wang2021IBRNetLM} fail to restore the clear structure of the stick. Our method produces a relatively complete shape, comparable to SIMPLI-8L\cite{Solovev2022SelfimprovingMI}.

Among all the evaluation metrics, our approach achieves comparable or better visual quality with offline approaches, and performs superior to online approaches. More qualitative comparison results on Shiny\cite{Solovev2022SelfimprovingMI} and IBRNet collected\cite{Wang2021IBRNetLM} are shown in Fig. \ref{fig_extra1}. 
Our method produces clear results that accurately captures view-dependent effects, albeit with slightly less sharpness compared to SIMPLI-8L\cite{Solovev2022SelfimprovingMI}. This trade-off is made in exchange for reduced time spent constructing the light field representations.
We also compare our results on Spaces\cite{Flynn2019DeepViewVS} dataset with the baseline methods listed in Table. \ref{tab_spaces}.
This figure shows that our results are still comparable with the those of DeepView\cite{Flynn2019DeepViewVS}.

\begin{figure}
    \centering
    \includegraphics[width=\columnwidth]{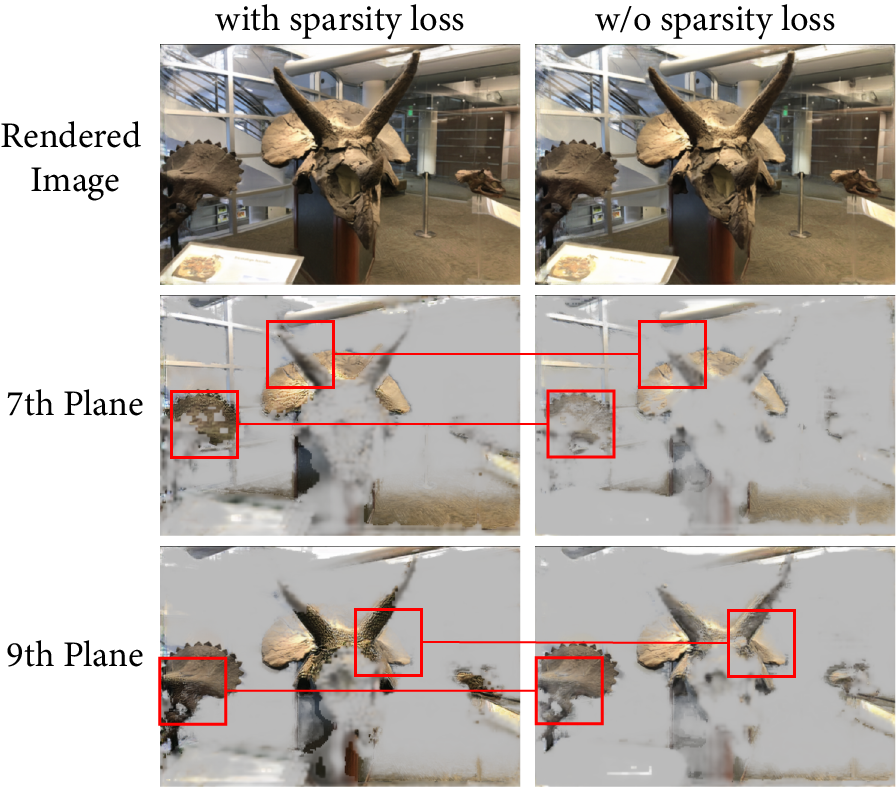}
    \caption{Qualitative comparison on the ablation configuration of \textit{with or without sparsity loss}. The left and right columns correspond to rendering results produced by the model trained with sparsity loss and without sparsity loss respectively. The PSNR for the left rendered image and the right rendered image are 28.51 and 27.42. The MPI planes trained with sparsity loss are clearer and less cloudy.}
    \label{fig_sparsity}
\end{figure}

\subsection{Ablation Study}
\label{sec_ablation}
In this section, we conduct a series of ablation experiments to assess the contribution of each component to the final rendering performance. The metrics of all experiments are evaluated on the Real Forward-facing\cite{Mildenhall2019LocalLF} dataset, which contains typical challenging cases such as occlusion, complex scene geometry, and view-dependent effects. The rendered images for training and evaluation are downscaled to $378\times 512$, and the number of input views is set to 3. The default number of planes in an MPI is 40. For detailed ablation settings and details, please refer to Tab. \ref{tab_ablation}.

\textbf{Sparsification factor $k$.} It is evident that increasing the value of $k$ results in enhanced rendering performance but at the expense of slower generation speed. Hence, it is crucial to identify an appropriate $k$ that strikes a balance between rendering quality and efficiency. Notably, as $k$ surpasses the value of 7, the demand for GPU memory during the training process exceeds the capacity of an RTX 3090 GPU. To circumvent this limitation, models with $k$ exceeding 7 have to be trained and inferred in smaller patches, which may inevitably downgrades the rendering quality. So we conducted ablation on $k$ with values 3, 5 and 7 in Tab. \ref{tab_ablation}. The results confirm that an increasing number of $k$ results in marginal increase in visual quality but obvious decrease in generation speed.  

\textbf{Number of iterations.} Comparing the configurations of the \textit{default model} with that of \textit{L=2 (2 iterations)} in Tab. \ref{tab_ablation}, it is evident that employing additional iterations of learned gradient descent yields only a marginal enhancement in rendering quality. However, this refinement results in a notable decrease of approximately 8 FPS in temporal efficiency. Therefore, to strike a balance between rendering quality and efficiency, employing a 2-iteration model suffices. This efficiency trade-off stems from the diminishing returns of increased iterations, where the incremental improvement in rendering quality does not adequately offset the cost in terms of temporal performance, hence motivating the preference for the 2-iteration model.

\textbf{Number of planes.} When comparing the configurations of the \textit{default model}, \textit{D=16 (fewer planes)}, and \textit{D=64 (more planes)} in Tab. \ref{tab_ablation}, it becomes evident that employing too few planes results in a substantial deterioration in rendering quality, albeit compensating for expedited generation speed. Conversely, an increase in the number of planes yields only marginal enhancements in rendering quality while significantly diminishing temporal efficiency. This observation suggests the presence of a potential saturation point for the number of MPI planes. Beyond this point, the incremental addition of planes may not yield linear improvements in performance.

\textbf{Sparsity Loss.} The \textit{default model} demonstrates superior rendering quality compared to the \textit{w/o sparsity loss} in Tab. \ref{tab_ablation}. And Fig. \ref{fig_sparsity} visually compares the differences in the rendered images and individual MPI planes from experiments conducted with and without the sparsity loss. Compared to the model trained with sparsity loss, the model trained without it generates rendered images that display more artifacts, particularly around sharp edges. A comparison of individual MPI planes shows that those produced by the model with sparsity loss contain clearer surfaces. This evidence suggests that sparsity loss aids the network in minimizing scene content layout ambiguity, resulting in cleaner and sharper images.

\textbf{Backbone network.} Unlike the most straightforward way to process the sparse MPI gradients with an \textit{MLP}, we employ a 3D CNN for better performance as discussed in Sec. \ref{method_update}. We can see from Tab. \ref{tab_ablation} that the configuration of \textit{MLP} backbone network results in a reduction in rendering quality when compared to the \textit{default model}. The diminished quality is attributed to the limited receptive field of an MLP. Given that the default 3D CNN structure aims to strike a balance between rendering quality and efficiency, further exploration of a more robust and efficient architecture is warranted to enhance performance.

\textbf{Spasification criterion.} In Section \ref{method_sparsify}, we referenced an MVS\cite{Chen2023UncertaintyAW,Jiang2023AdaptMVSNetEM,Weilharter2021HighResMVSNetAF,360MVSNet,Uncertainty,Multistage}-style gradient sparsification approach that involves selecting the layer with the highest alpha gradient along the depth axis as well as its adjacent $(k - 1)/2$ layers to form a sparsified volume of $k$ layers of gradient data. This strategy, however, only takes geometry into consideration and may not adequately capture certain light field intricacies, such as semi-transparent objects and thin structures. An overall comparison in Table \ref{tab_ablation} reveals that the \textit{MVS-sampling} strategy exhibits lower rendering quality when compared with the \textit{default model}. 

\textbf{Full-sized inference without HSGD.} Leveraging our proposed Hierarchical Sparse Gradient Descent (HSGD) strategy, we are able to generate a full-sized MPI in a single forward pass. To examine the impact of this design, we also evaluated a patch-based inference configuration similar to DeepView: training and inferring on smaller patch-sized MPIs, cropping the edges of each patch, and merging them to form a complete full-sized MPI. For this comparison, we disabled the HSGD strategy, inferred four patches per full-sized MPI, and merged them into one. As shown in Table \ref{tab_ablation}, the “w/o HSGD” configuration exhibits a significant drop in efficiency, as it requires four forward passes and additional processing to crop and merge the patches. Moreover, these steps fail to completely eliminate visual discontinuities between patches, resulting in noticeable degradation in both efficiency and quality.

\begin{table}[htbp]
\caption{Quantitative ablation of the design choice on Real Forward-facing evaluation set.}
\centering
\resizebox{\columnwidth}{!}{
\begin{tabular}{*{1}{l}*{4}{c}}
\toprule
Configuration & \makecell{Generation\\ FPS$\uparrow$} & PSNR $\uparrow$ & SSIM$\uparrow$ & LPIPS$\downarrow$ \\%
\midrule
$k = 3$ & 43.29 & 24.95 & 0.83 & 0.17 \\
$k = 7$ & 32.79 & \textbf{25.51} & \textbf{0.84} & \textbf{0.15} \\
$D = 16$(fewer planes) & \textbf{54.64} & 24.39 & 0.79 & 0.19 \\
$D = 64$(more planes) & 28.57 & 25.48 & 0.84 & 0.15 \\
$L = 2$(2 iterartions) & 45.45 & 25.33 & 0.84 & 0.16 \\
w/o sparsity loss & 31.73 & 25.20 & 0.81 & 0.16 \\
MLP & 48.54 & 24.86 & 0.82 & 0.17 \\
MVS-sampling & 37.04 & 24.34 & 0.79 & 0.19 \\
w/o HSGD & 0.52 & 23.78 & 0.71 & 0.23 \\
\midrule
default model & 37.04 & 25.46 & 0.84 & 0.15 \\
\bottomrule
\end{tabular}
}
\label{tab_ablation}
\end{table}
\section{Conclusion}
This paper presents a novel method for efficiently reconstructing light fields using hierarchical sparse gradient descent. Our proposed method achieves a resolution of $378\times 512$ for light field representations (MPIs) at a frame rate of above 35 FPS, from which novel views can be rendered at around 700 FPS. The hierarchical sparse gradient descent module of our network focuses on scene-aligned sparse MPI gradients, resulting in significant improvements in temporal efficiency without compromising rendering quality. Our method has the potential to deliver high-quality and real-time light field videos for XR and Naked Eye 3D display devices.

\begin{figure}[!t]
  \centering
  \includegraphics[width=\columnwidth]{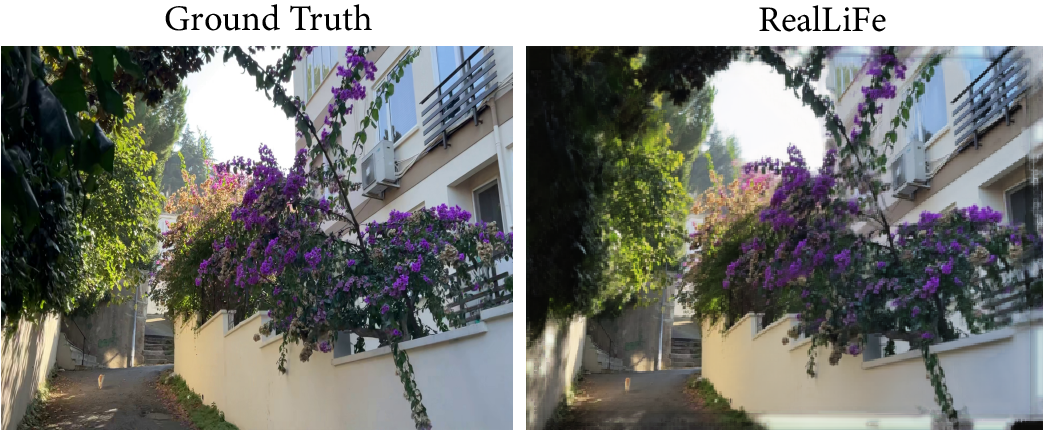}
  \caption{The border artifacts of our approach. The borders of our rendered image contains obvious color difference.}
  \label{fig_border}
\end{figure}

\textbf{Limitations.} There are some limitations of our approach. Firstly, our current implementation can only be trained with a fixed number of views, and the order of input views may slightly affect the final rendering quality. Secondly, our model is unable to produce relatively good results at the borders of the image where not all source views are overlapped, shown in Fig. \ref{fig_border}. 
Thirdly, large baselines pose an inherent challenge for discrete-depth MPI methods including ours because the disparity range grows nonlinearly, causing scene surfaces to fall between sampled depth planes and producing ambiguous or conflicting evidence in the plane-sweep volume (PSV). Resolving these ambiguities requires long-range visibility reasoning across many depth planes, which deeper networks such as DeepView naturally support but lightweight backbones struggle to capture. In our case, this leads to reduced reconstruction stability under extreme viewpoint changes, particularly when PSV overlap becomes small. Increasing depth-plane density or adopting a larger backbone can mitigate the issue but at a significant computational cost. Developing stronger geometric or generative priors for handling PSV ambiguity remains a promising direction for improving robustness under large baselines.

\textbf{Future work.} Although the proposed approach is able to generate light fields at above 35 FPS of resolution $378\times 512$, its efficiency could still be improved with customized CUDA kernels, which is not emphasized in this paper. Furthermore, a robust and adaptive view-aggregating module is required to support an arbitrary number of input views. This enhancement will facilitate the use of our model for various camera configurations.
\section{Acknowledgements}
This work is supported in part by Natural Science Foundation of China (NSFC) under contract No.62125106 and 62427804, in part by Tsinghua University Dushi Program (No.20251080107), in part by Beijing Outstanding Young Scientist Program under contract No.JWZQ20240101009, and in part by the Xplorer Prize.

\bibliography{citations}
\bibliographystyle{IEEEtran}

\begin{IEEEbiography}
[{\includegraphics[width=1in,height=1.25in,clip,keepaspectratio]{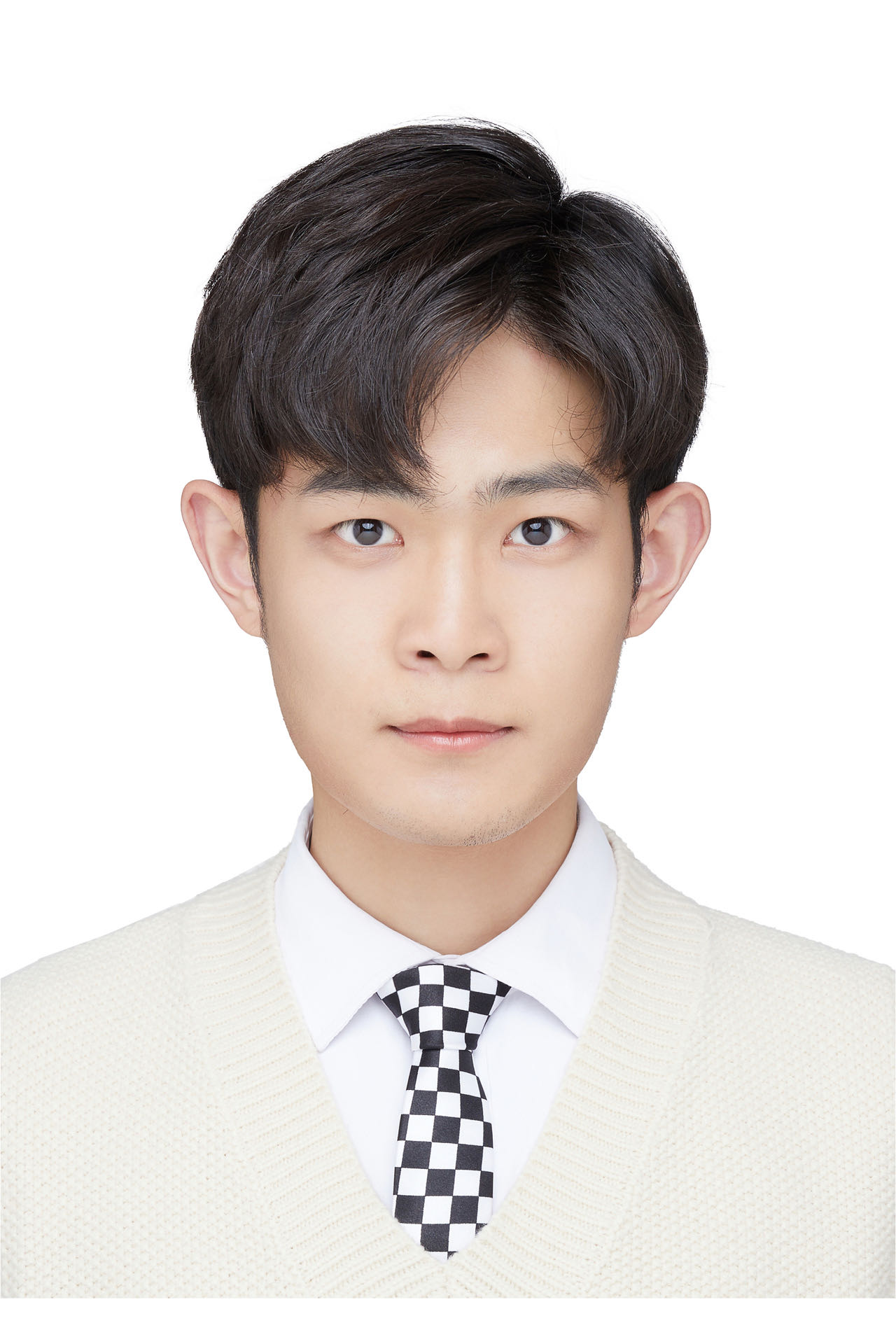}}]{Yijie Deng} is currently a PhD candidate student in New York University. He received Master's degree in Tsinghua-Berkeley Shenzhen Institute (TBSI), Tsinghua University in 2024. He received B.E. from Wuhan University in 2021. His research interest is 3D vision, Graphics and Robotics.
\end{IEEEbiography}

\begin{IEEEbiography}
[{\includegraphics[width=1in,height=1.25in,clip,keepaspectratio]{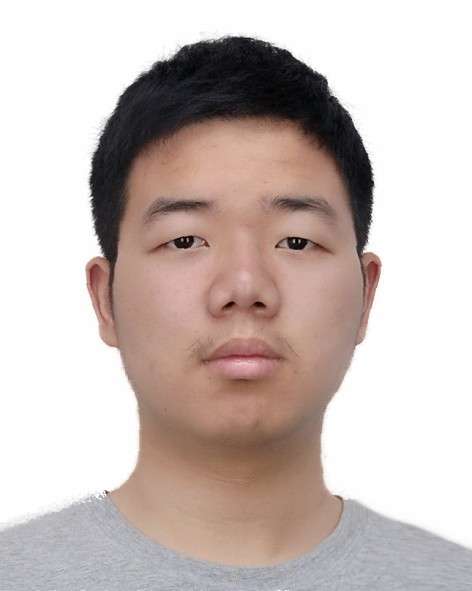}}]{Lei Han} is currently a researcher in Tsinghua University. He studied Electrical Engineering at the Hong Kong University of Science and Technology and Tsinghua University. He received the B.S. degree in July 2013 and joined the Department of Electrical Computing Engineering at the Hong Kong University of Science and Technology in September 2016, where he is pursuing the PhD degree. His current research focuses on multi-view geometry and 3D computer vision.
\end{IEEEbiography}

\begin{IEEEbiography}
[{\includegraphics[width=1in,height=1.25in,clip,keepaspectratio]{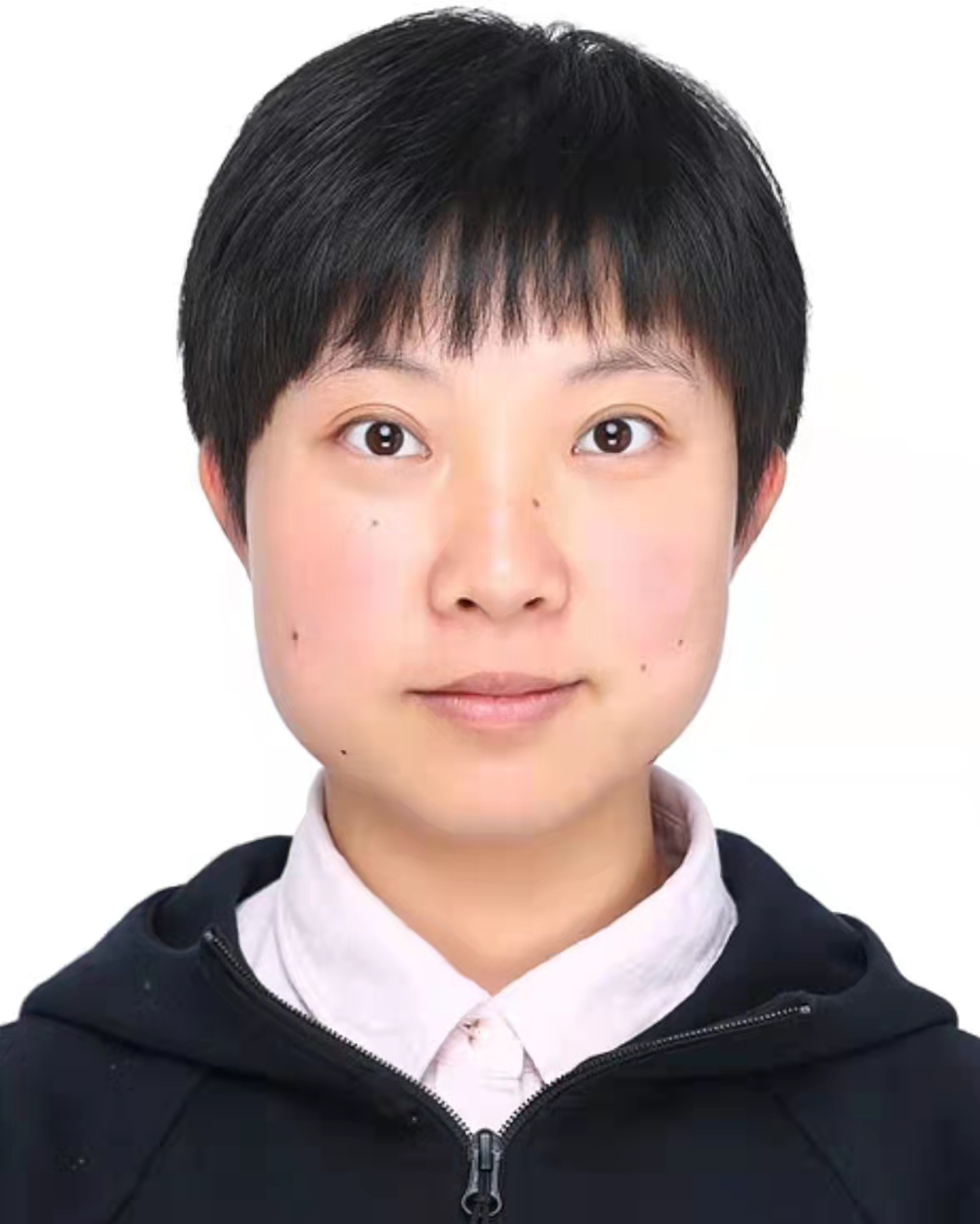}}]{Lin Li} is currently an engineer in Huawei. She received the B.S. and the integrated M.S. and Ph. D. degrees from the Department of Electronic Information Technology and Instrument Institute of ZheJiang University in 2010 and 2015, respectively.
\end{IEEEbiography}

\begin{IEEEbiography}
[{\includegraphics[width=1in,height=1.25in,clip,keepaspectratio]{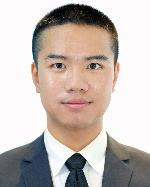}}]{Tianpeng Lin} researcher in Kirin Chipset and Linx Lab of HiSilicon (subsidiary of Huawei Technologies) from 2016 to 2023. He received Ph.D in GeoInformatics in the Chinese University of Hong Kong in 2016, and B.E. from Tsinghua University in 2009. He works for Kirin chipsets and Linx Lab and focuses on the chipset algorithm development in SLAM and 3D reconstruction.
\end{IEEEbiography}

\begin{IEEEbiography}[{\includegraphics[width=1in,height=1.25in,clip,keepaspectratio]{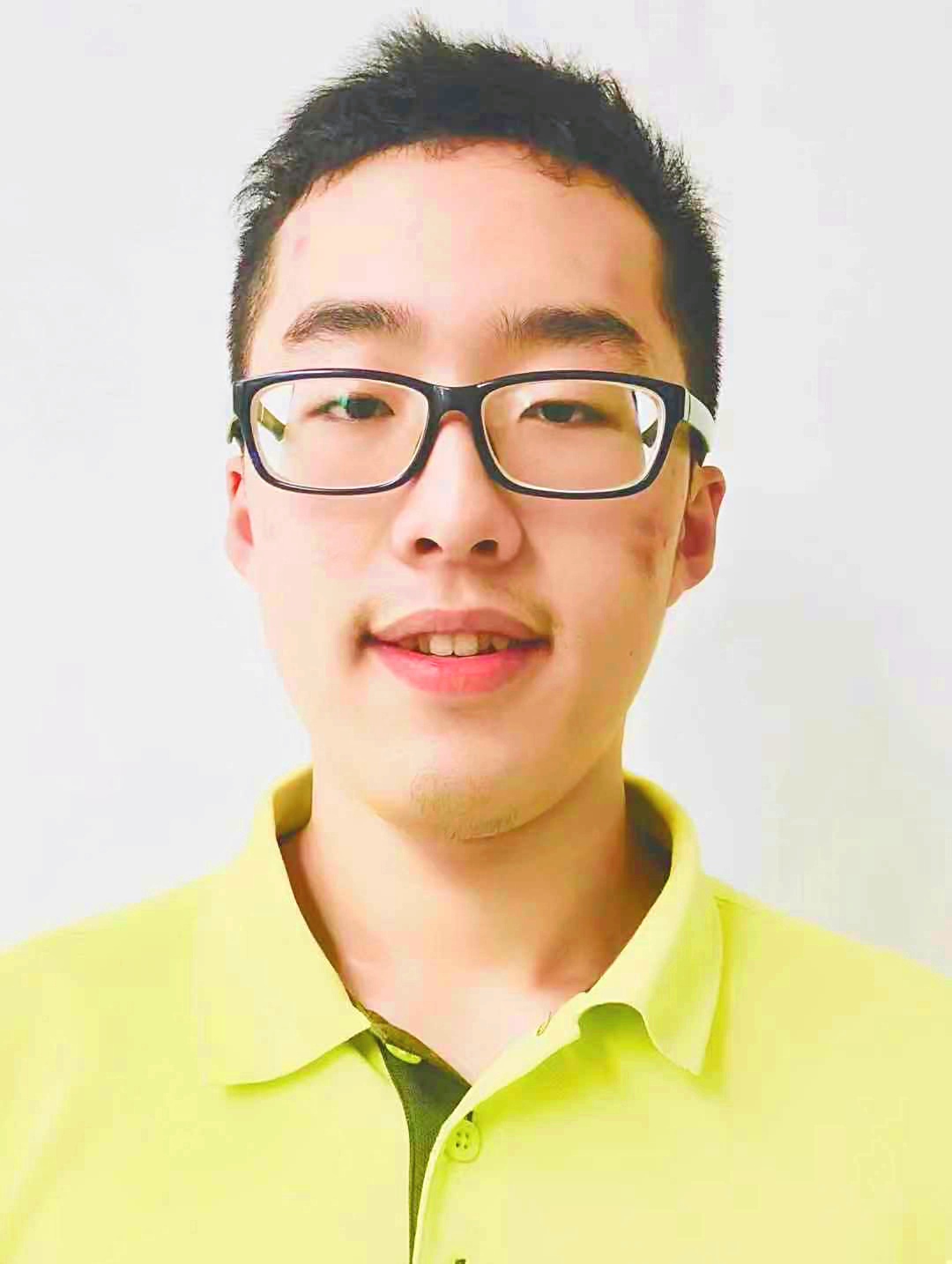}}]{Jinzhi Zhang} is currently a Ph.D student in Tsinghua-Berkeley Shenzhen Institute (TBSI), Tsinghua University. He received B.E. from Huazhong University of Science and Technology in 2019. His research interest is 3D vision.
\end{IEEEbiography}

\begin{IEEEbiography}[{\includegraphics[width=1in,height=1.25in,clip,keepaspectratio]{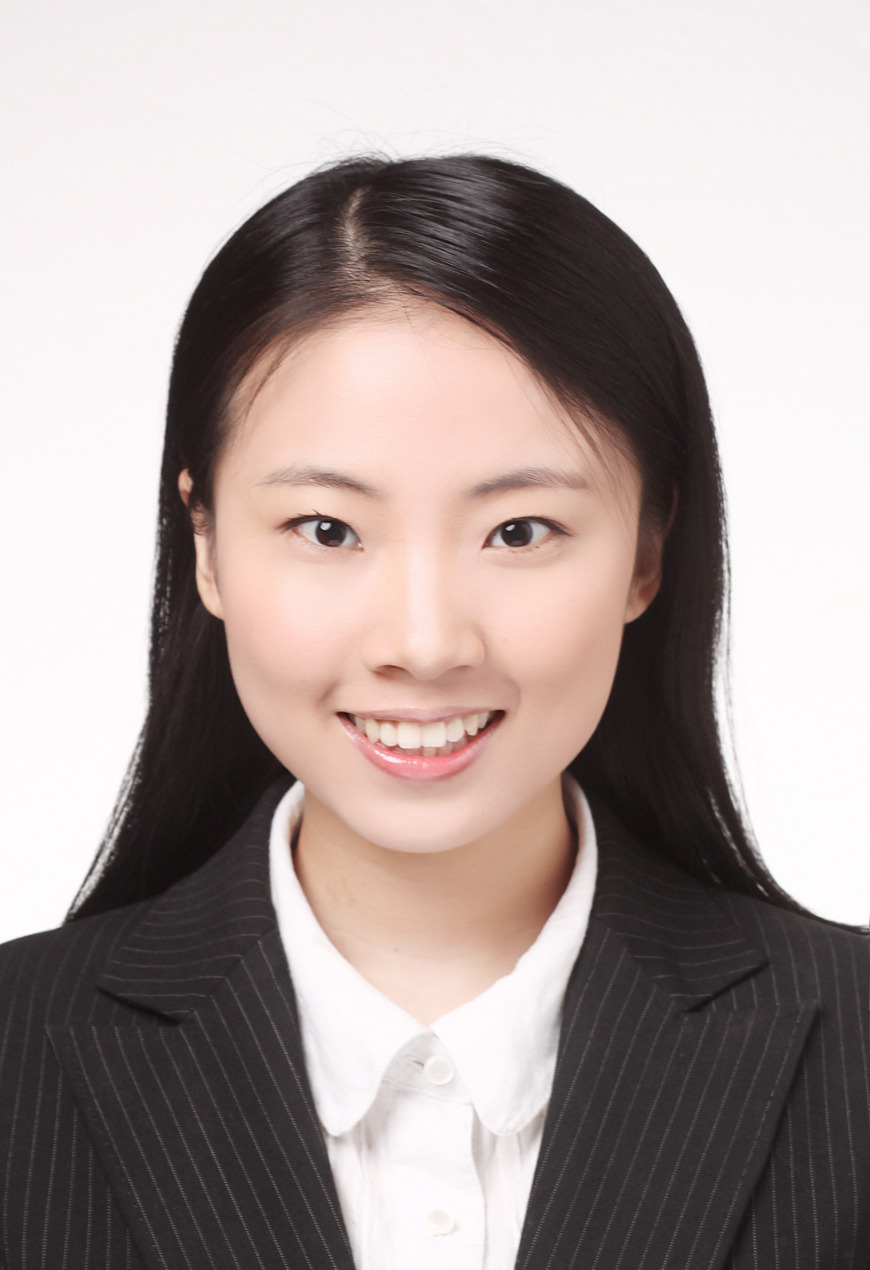}}]{Dr. Fang} is a Professor in the Department of Electronic Engineering at Tsinghua University. Her research integrates physical optics with artificial intelligence to advance next-generation imaging and computing technologies. She has published over 100 peer-reviewed articles in journals such as Science, Nature, Nature Photonics, and Nature Nanotechnology. Dr. Fang is a member of the IEEE MMSP Technical Committee, an Associate Editor for IEEE TIP and Optica, and serves as the General Chair of IEEE MMSP 2025.
\end{IEEEbiography}

\end{document}